\documentclass[sigconf]{acmart}
\renewcommand\footnotetextcopyrightpermission[1]{}
\setcopyright{none}

\makeatletter
\let\oldfnsymbol\@fnsymbol
\renewcommand{\@fnsymbol}[1]{%
  \ifnum#1=2\relax
    \mbox{\ding{41}}%
  \else
    \oldfnsymbol{#1}%
  \fi
}
\makeatother

\usepackage{float}
\usepackage{multirow}
\AtBeginDocument{%
  }

\begin{document}

\title{Skill2Query: Exploiting Skill Structure to Generate Pseudo-Queries for Agent Skill Retrieval}


\author{Lihui Ding}
\authornote{These authors contributed equally to this work.}
\affiliation{%
  \institution{Fudan University}
  \city{Shanghai}
  \country{China}
}
\email{dinglh25@m.fudan.edu.cn}

\author{Zihan Guo}
\authornotemark[1]
\authornote{Corresponding authors.}
\affiliation{%
  \institution{Sun Yat-sen University}
  \city{Shenzhen}
  \country{China}
}
\affiliation{%
  \institution{Shanghai Innovation Institute}
  \city{Shanghai}
  \country{China}
}
\email{guozh29@mail2.sysu.edu.cn}

\author{Bingwei Lu}
\affiliation{%
  \institution{Shanghai Jiao Tong University}
  \city{Shanghai}
  \country{China}
}
\email{icestring@sjtu.edu.cn}

\author{Chenyu Zhou}
\affiliation{%
  \institution{Shanghai Jiao Tong University}
  \city{Shanghai}
  \country{China}
}
\email{chenyuzhou@sjtu.edu.cn}

\author{Yuanjian Zhou}
\affiliation{%
  \institution{Shanghai Innovation Institute}
  \city{Shanghai}
  \country{China}
}
\email{jake.zhou@sii.edu.cn}

\author{Weinan Zhang}
\affiliation{%
  \institution{Shanghai Jiao Tong University}
  \city{Shanghai}
  \country{China}
}
\affiliation{%
  \institution{Shanghai Innovation Institute}
  \city{Shanghai}
  \country{China}
}
\email{wnzhang@sjtu.edu.cn}

\author{Jianghao Lin}
\authornotemark[2]
\affiliation{%
  \institution{Shanghai Jiao Tong University}
  \city{Shanghai}
  \country{China}
}
\email{linjianghao@sjtu.edu.cn}

\author{Dongdong Ge}
\affiliation{%
  \institution{Shanghai Jiao Tong University}
  \city{Shanghai}
  \country{China}
}
\email{ddge@sjtu.edu.cn}

\renewcommand{\shortauthors}{Ding et al.}
\begin{abstract}
Pseudo-query generation can alleviate the supervision bottleneck for agent skill retrieval, but existing document-level approaches typically leave the rich internal relations among capabilities, parameters, and usage examples implicit. As a result, generated queries may be topically relevant to a skill while lacking capability grounding and parameter consistency, raising the question of whether explicitly exploiting a skill document’s internal structure can produce more effective retrieval signals. We therefore propose Skill2Query, a framework that first parses a skill document into a Skill Knowledge Graph and then generates pseudo‑queries through a three‑stage process including style mimicking, query template generation, and parameter filling. The generated queries can be used for offline index augmentation, online query expansion, and retriever training. Four benchmarks (TheoremQA, LogicBench, ToolQA, and CHAMP) are used to evaluate Skill2Query with large-scale skill candidate pools across multiple downstream applications, including skill retrieval, retriever training, and end-to-end agent execution. Using nearly 30K skills across diverse domains, we generate 700K category-diverse pseudo-queries. Skill2Query consistently improves sparse, dense, and skill-routing retrieval, with an average Recall@1 gain of 6.70 percentage points across retrieval settings. Skill2Query-generated training data also achieves the best Recall@1 and nDCG@1 among the evaluated generation baselines. Further evaluations with multiple LLM backends demonstrate that improved skill retrieval translates into higher agent task success rates.Code and resources are available at \url{https://github.com/MatZaharia/Skill2Query}.
\end{abstract}


\maketitle
\pagestyle{plain}

\section{Introduction}

Modern AI agents extend their capabilities by retrieving and invoking reusable skills~\cite{patil2024gorilla,qin2024toolllm,li2023api,zhou2026externalization,guo2026skillprobe}. 
As the number of available capabilities grows from a small manually curated set to repositories containing thousands of candidates, an agent must first identify the skills relevant to a user request before it can plan or execute the task~\cite{li2026skillflow,zheng2026skillrouter}. 
Retrieval errors at this stage may expose the agent to irrelevant capabilities or exclude the skill required for task completion~\cite{shi2025retrieval}.
Accurate skill retrieval has therefore become an important component of large-scale agent systems.


Skill retrieval presents an asymmetric matching problem. 
Users typically describe what they want to accomplish through short, colloquial, and sometimes underspecified queries~\cite{lin2025masstool}. 
Skill documents, in contrast, are written by developers to specify what a skill provides, often using technical descriptions, parameter declarations, constraints, and usage examples. 
Established retrieval architectures can match lexical or semantic similarity between the two sides~\cite{robertson2009probabilistic,reimers2019sentence,xiao2024c}, but their effectiveness depends on access to high-quality query--skill supervision. 
Constructing such supervision manually requires annotators to distinguish among many semantically overlapping skills and understand their functional boundaries, making large-scale annotation expensive~\cite{bonifacio2022inpars,wang2022gpl,dai2022promptagator}.


Pseudo-query generation offers a scalable way to alleviate this supervision bottleneck~\cite{nogueira2019document,bonifacio2022inpars}. 
Methods such as doc2query and InPars generate synthetic queries from target documents and use them for document expansion or retriever training.
When adapted to skill retrieval, however, many document-level pseudo-query methods condition generation on a skill name, description, or full document represented as a single text sequence.
This document-level generation objective encourages topical relevance to the skill as a whole, while leaving the document's internal structure implicit.
A skill document may describe several distinct capabilities, associate them with different parameters, and provide examples for only particular usage patterns.
Consequently, a generated query can be semantically related to the skill without being grounded in a specific capability or using parameter values consistent with the supported interface. 
The limitation is therefore not whether the generator has access to the full skill document, but whether it explicitly exploits the structure contained in that document.


Taken together, existing pseudo-query methods are designed to reduce the expression mismatch between user queries and skill documents, but they leave a skill-structure utilization gap that has received limited attention. 
Skill documents contain rich internal structure, whereas existing generation processes typically model this information implicitly as part of a complete document rather than turning it into explicit generation constraints~\cite{nogueira2019document,bonifacio2022inpars,
wang2022gpl,dai2022promptagator}. 
This gap has two main manifestations.
First, the capability-grounding gap arises when a generated query is semantically related to the skill as a whole but is not associated with a specific capability. 
Second, the parameter-consistency gap arises when the parameters or parameter values expressed in a query do not conform to the interface and constraints of the corresponding capability. 
Based on this observation, we investigate whether explicitly exploiting the internal structure of skill documents can generate pseudo-queries with finer-grained functional grounding and stronger parameter consistency, thereby providing more effective retrieval signals for agent skill retrieval.


Based on this hypothesis, we propose Skill2Query, a framework that exploits skill structure to generate pseudo-queries for agent skill retrieval.
Skill2Query first parses each skill document into a Skill Knowledge Graph (SKG) that represents its capabilities, parameters, examples, and their relationships.
It then applies a three-stage generation process.
A Style Mimicker extracts expression patterns from usage examples, a Query Template Generator associates query templates with individual capabilities and parameter slots, and a Param Filler instantiates and validates the slots according to the parameter definitions.
The resulting pseudo-queries can be used in three settings, including offline index augmentation, online query expansion, and query--skill supervision for retriever training.


The contributions of this work are threefold. 
1) We formulate the underexplored skill-structure utilization gap in
pseudo-query generation for agent skill retrieval, characterizing it in terms of capability grounding and parameter consistency.
2) We propose Skill2Query, which constructs a Skill Knowledge Graph and generates pseudo-queries through style extraction, query template generation, and parameter filling, supporting index augmentation, query expansion, and retriever training. 
3) Experiments on four datasets across three retrieval paradigms demonstrate improvements in both pseudo-query quality and retrieval effectiveness, and an end-to-end case study further indicates potential task-level utility.

\section{Related Work}
Our work intersects three research directions: agent skill retrieval and invocation, pseudo-query generation and query expansion, and structure-aware skill representation for agents.

\subsection{Agent Skill Retrieval and Invocation}

As large language models evolve from general-purpose text generators into agents capable of invoking external tools and skills~\cite{wang2026skills}, accurately selecting the executable skills required for a task from a large-scale skill library has become a critical problem in agent systems~\cite{li2026skillflow,cho2026skillret}. Unlike traditional tool invocation, agent skills typically include not only brief functional descriptions but also execution steps, parameter specifications, constraints, examples, and contextual usage patterns~\cite{pan2026skillmas}. 
Skill retrieval is therefore not merely a semantic matching problem.
It also requires precise correspondences among user intent, skill capability boundaries, and executable conditions.

Early tool retrieval research has laid important foundations for skill retrieval. 
Gorilla demonstrated the effectiveness of retrieval augmentation in improving API invocation accuracy by retrieving relevant APIs from a large-scale API documentation corpus~\cite{patil2024gorilla}. 
ToolLLM and API-Bank further evaluated LLM tool-use capabilities across multiple
dimensions, including tool retrieval, tool invocation, and tool
planning~\cite{qin2024toolllm,li2023api}. These works indicate that retrieval
quality directly affects an agent's ability to invoke external capabilities.
Compared with concise API specifications, agent skill documents may contain longer and more heterogeneous combinations of functional descriptions, execution procedures, constraints, parameters, and examples, creating additional representation and supervision challenges~\cite{kachuee2025improving}.

In the agent skill setting, SkillFlow is among the earliest systems designed
for retrieving community-contributed skill documents. The system indexes
approximately 36K community-contributed \texttt{SKILL.md} files and employs a
cascaded architecture consisting of dense retrieval, cross-encoder reranking,
and LLM-based selection~\cite{li2026skillflow}. A key finding of
SkillFlow is that exposing only a skill's name and description while concealing
the full body significantly degrades routing accuracy, indicating that the full
skill text is a critical semantic signal in skill retrieval. 
SkillRouter further validates this conclusion at a larger scale of approximately 80K candidate skills, noting that existing progressive disclosure designs can severely impair retrieval effectiveness under
large-scale semantic overlap~\cite{zheng2026skillrouter}. 

These works collectively demonstrate that skill retrieval performance is
highly dependent on the complete and fine-grained information contained in
skill documents, particularly the functional scope, parameter requirements,
and execution constraints implicitly encoded in the skill body. However,
existing research has primarily focused on retrieval architecture (for
example, how to design recall, reranking, and cascaded selection
pipelines) without systematically addressing the origin of high-quality
query--skill training data.

\subsection{Pseudo-Query Generation and Synthetic Supervision}
Pseudo-query generation aims to automatically construct synthetic queries
relevant to target documents in order to mitigate the representation gap
between queries and documents and provide training supervision for retrieval
models. Classical methods such as doc2query append generated queries to
documents to enhance retrieval indices~\cite{nogueira2019document,chen2408re}, while
InPars leverages large language models to generate query--document pairs for
unsupervised retriever training~\cite{bonifacio2022inpars}. HyDE and Query2doc
generate hypothetical documents or pseudo-documents from the query side,
further demonstrating that generative text can effectively bridge the semantic
gap between queries and retrieval targets
~\cite{gao2023precise,wang2023query2doc}.

In the context of tool/API retrieval and agent tool learning, synthetic
supervision has also been employed to alleviate the scarcity of human
annotations. Related methods typically generate queries directly from tool
names, functional descriptions, or a small number of examples, or construct
training data through document expansion and query rewriting. Seal-Tools
constructs tool-learning datasets via self-instruct
\cite{wang2023self,wu2024seal}, while recent work investigates whether LLM-generated
synthetic query rewrites can better capture user intent in
retrieval-augmented generation
\cite{zheng2025can}. These works indicate that synthetic data can provide
effective supervision for retrieval and tool learning~\cite{jiang2023noisy}.

However, most existing methods treat input documents or tool descriptions as unstructured text to be rewritten or expanded. 
While this assumption is acceptable for general document retrieval or simple tool/API retrieval, it causes critical structural signal loss in skill retrieval. Skill documents typically contain parameter schemas, type constraints, capability boundaries, and usage examples.
When the document is provided only as flat text, the generator is not explicitly constrained by capability--parameter relations and therefore provides no deterministic guarantee of parameter consistency.

\subsection{Structure-Aware Skill Representation and Retrieval}
Structure-aware skill representation aims to go beyond plain-text
descriptions by organizing agent capabilities through explicit semantic
structures, while recent evidence suggests that skill organization itself can
affect agent behavior~\cite{chen2026skilljuror}. Alternative skill representations have also been explored in weight
space~\cite{yu2026latentskill}. Graph-of-Skills constructs a dependency-aware skill graph for
large-scale agent skills and combines seed retrieval, reverse-aware
Personalized PageRank, and contextual budget constraints to retrieve more
compact and dependency-complete skill bundles
\cite{liu2026graph}. This method primarily models inter-skill
relationships, including dependencies, workflows, semantic similarity, and
substitution relations, with the goal of recovering the skill set required
for a task at inference time. In contrast, Skill2Query focuses on intra-skill structural relationships, including metadata, capabilities, parameters, and examples, and transforms them into parameter-aware pseudo-queries for index augmentation and retrieval model training.

The API and tool construction domain has also validated the importance of
structured representation. Let's Chat to Find the APIs leverages knowledge
graphs to organize API capabilities and combines structured API knowledge with
LLM reasoning~\cite{huang2023let}. OpenAPI provides formalized descriptions of
API endpoints, parameter types, request schemas, and invocation
formats~\cite{openapi2021specification}. ToolFactory automatically extracts
structured API information from unstructured REST API documentation and
converts it into tools callable by agents~\cite{ni2025toolfactory}. These works
demonstrate that structured representations contribute to improved tool
discovery, tool construction, and invocation reliability.

In summary, prior work has studied retrieval architectures, synthetic supervision, and structured tool representations largely as separate problems, while relatively little work has examined how intra-skill structure can be converted directly into pseudo-query supervision.
Skill2Query operates precisely at this intersection: it parses
intra-skill semantic structures via SKG and generates parameter-aware
pseudo-queries for offline index augmentation, online query expansion, and retriever training.

\section{Method}
\begin{figure*}[t]
{
\centering
\includegraphics[width=\textwidth]{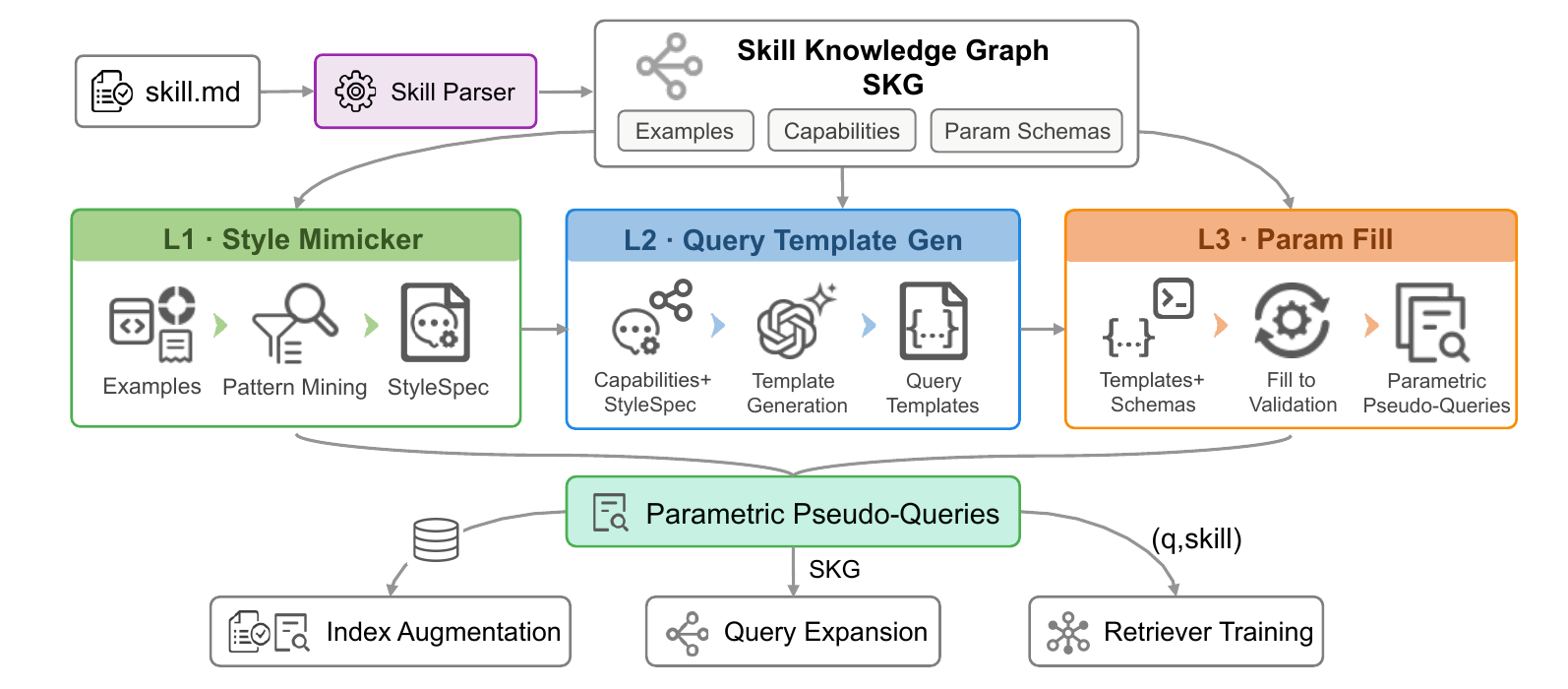}
}
\caption{Overview of the Skill2Query framework.}
\Description{The figure presents the overall workflow of Skill2Query, including Skill Knowledge Graph construction, three-stage progressive pseudo-query generation, offline index construction, and online retrieval with query expansion.}
\label{fig:framework}
\end{figure*}

Skill2Query uses the SKG as a structured intermediate representation and applies a three-stage pipeline to convert skill documents into capability- and parameter-aware pseudo-queries.
Figure~\ref{fig:framework} provides an overview of the proposed Skill2Query framework.

\subsection{Problem Formalization}
Consider a skill repository
\[
\mathcal{S} = \{s_1, s_2, \ldots, s_N\},
\]
where each skill \(s_i\) is defined by a \texttt{SKILL.md} document containing the skill name, functional description, parameter declarations, usage examples, and other metadata.

Given a user's natural-language query \(q \in \mathcal{Q}\), the goal of the retrieval system is to return a subset of skills from \(\mathcal{S}\) that are semantically matched to \(q\) in terms of both functional compatibility and consistency between query-implied arguments and the skill interface. Formally, we define the retrieval problem as finding a mapping
\[
R : \mathcal{Q} \rightarrow 2^{\mathcal{S}},
\]
such that, for any query \(q \in \mathcal{Q}\), the returned skill set
\(R(q)\) satisfies:

\begin{enumerate}
    \item semantic compatibility: the task intent expressed by the
    user query is consistent with the functionality of the returned skills.

    \item parameter compatibility: the input conditions and
    constraints conveyed by the user query are consistent with the parameter definitions supported by the returned skills.
\end{enumerate}

The core idea of Skill2Query is as follows. For each skill \(s_i\), we first
generate a set of query templates containing parameter placeholders based on
the structured information extracted from the corresponding skill document.
These templates are then instantiated with concrete parameter values to
produce a pseudo-query set
\[
\widetilde{\mathcal{Q}}_i
=
\left\{
\widetilde{q}_{i,1},
\widetilde{q}_{i,2},
\ldots,
\widetilde{q}_{i,M_i}
\right\}.
\]

The pseudo-query set is constructed such that, for any real user query \(q \in \mathcal{Q}\), if the intent of \(q\) matches skill \(s_i\), there exists at least one pseudo-query \(\widetilde{q}_{i,j} \in \widetilde{\mathcal{Q}}_i\) that has high similarity to \(q\) in the semantic space. 
Accordingly, the retrieval process can be reduced to:
\[
R_k(q)
=
\operatorname{TopK}_{s_i \in \mathcal{S}}
f\left(q, s_i, \widetilde{\mathcal{Q}}_i\right),
\]
where \(f\) denotes the scoring function of the underlying retrieval configuration. It may correspond to lexical matching, dense similarity, or a learned reranking score, while \(\widetilde{\mathcal{Q}}_i\) augments the representation associated with \(s_i\).


\subsection{Skill Knowledge Graph}

To systematically organize the information of skills, we construct a
Skill Knowledge Graph
\[
G_i = \left(V_i, A_i, T_V, T_E\right)
\]
for each skill \(s_i\), where \(V_i\) is the set of nodes, \(A_i\) is the set
of edges, and \(T_V\) and \(T_E\) are the sets of node types and edge types,
respectively.

\subsubsection{Node Types}

The SKG defines four categories of nodes:
\[
T_V
=
\{
\text{Skill},
\text{Capability},
\text{Parameter},
\text{Example}
\}.
\]

Skill node \(v_s\) represents the skill itself, with attributes defined as
\[
\operatorname{attr}(v_s)
=
(
\text{skill\_id},
\text{name},
\text{description},
\text{body}
),
\]
where \texttt{skill\_id} denotes the unique identifier of the skill,
\texttt{name} represents the skill name, \texttt{description} refers to the
functional description, and \texttt{body} contains the complete skill
document.

The set of Capability nodes associated with skill \(s_i\) is defined as
\[
C_i
=
\{
v_{c,1},
\ldots,
v_{c,m_i}
\},
\]
where \(m_i\) denotes the number of capabilities provided by the skill.

Each
Capability node represents an independent function, with attributes defined as
\[
\operatorname{attr}(v_{c,k})
=
(
\text{capability\_id},
\text{text}
).
\]

By explicitly modeling Capabilities as independent nodes, the query template
generator can generate queries around different functional aspects, thereby
improving functional coverage.

The set of Parameter nodes associated with skill \(s_i\) is defined as
\[
P_i
=
\{
v_{p,1},
\ldots,
v_{p,r_i}
\},
\]
where \(r_i\) denotes the number of parameters in the skill.

Each Parameter node contains the parameter information:
\[
\Pi_{i,j}
=
(
\text{name},
\text{type},
\text{required},
\text{default},
\text{min},
\text{max},
\text{enum},
\text{examples}
),
\]
where \texttt{name}, \texttt{type}, \texttt{required}, and
\texttt{default} represent the parameter name, data type, requirement status,
and default value, respectively. \texttt{min} and \texttt{max} define the
valid range, \texttt{enum} specifies enumeration constraints, and
\texttt{examples} provides typical parameter values extracted from the skill
document.

The set of Example nodes associated with skill \(s_i\) is defined as
\[
E_i
=
\{
v_{e,1},
\ldots,
v_{e,n_i}
\},
\]
where \(n_i\) denotes the number of examples contained in the skill document.

Each Example node represents an official usage example, with attributes
defined as
\[
\operatorname{attr}(v_{e,j})
=
(
\text{query},
\phi_{i,j}
),
\]
where \texttt{query} represents the natural language query in the example, and
\(\phi_{i,j}\) denotes the parameters and corresponding values appearing in
the example. 

\subsubsection{Edge Types}

The five types of directed edges form the edge type set
\[
T_E
=
\{
e_{\mathrm{has\_capability}},
e_{\mathrm{has\_parameter}},
e_{\mathrm{has\_example}},
e_{\mathrm{fill\_param}},
e_{\mathrm{demonstrates}}
\},
\]
which is used to characterize relationships among different types of
nodes. 

\begin{figure}[t]
    \centering
    \includegraphics[width=\columnwidth]{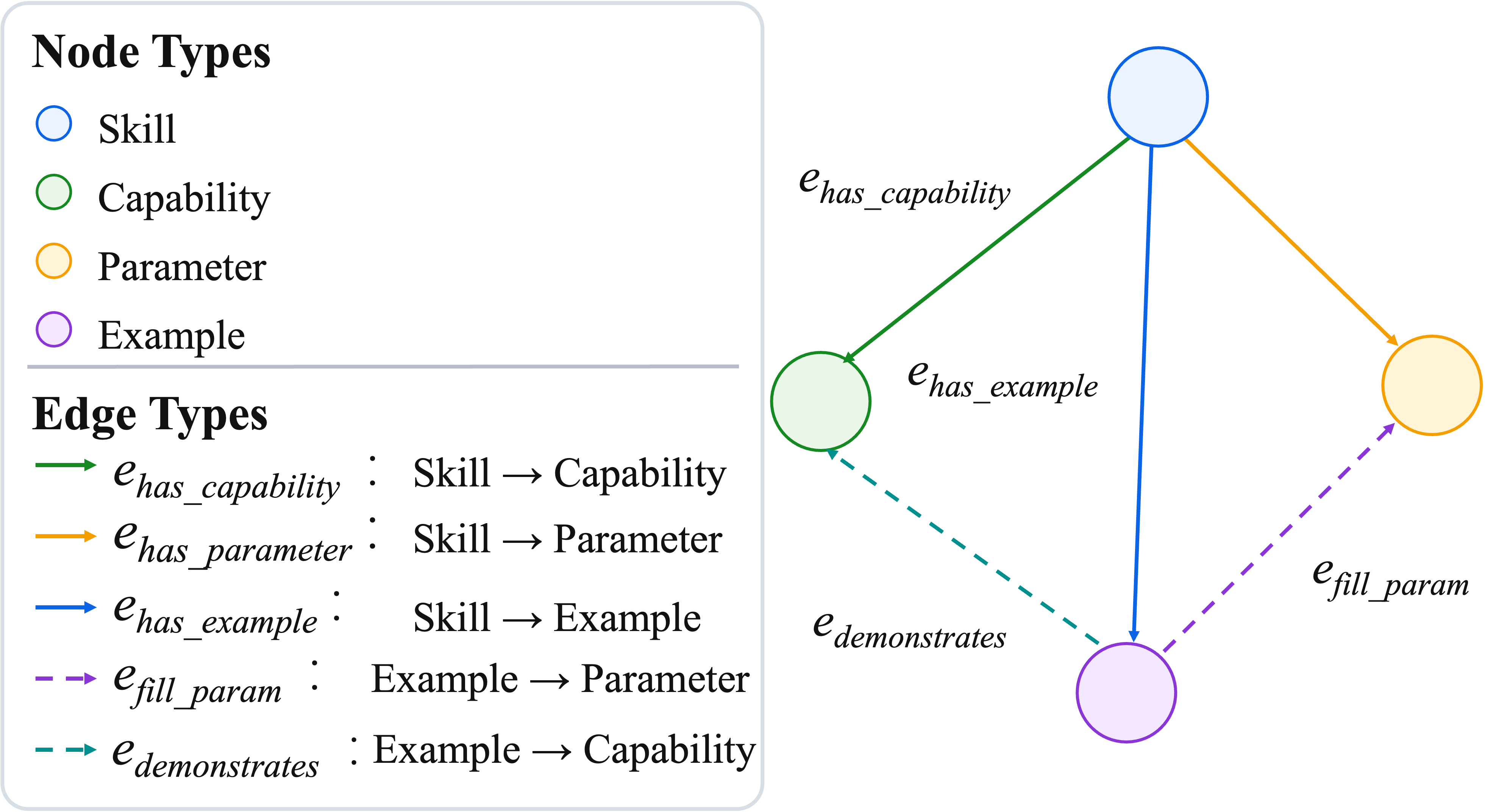}
    \caption{The node and edge types defined in the SKG.
    Solid arrows start from Skill nodes, and dashed arrows start from Example nodes.}
    \Description{The figure illustrates the node categories and edge types in SKG, including Skill, Capability, Parameter, and Example nodes, as well as the semantic relationships among them.}
    \label{fig:skg}
\end{figure}

As illustrated in Figure~\ref{fig:skg}, through the above nodes and edges, SKG
organizes the capabilities, parameters, and examples contained in skill
documents into an explicitly connected structure, providing a unified
representation for subsequent style extraction, query template generation, and parameter instantiation.

\subsection{Three-stage Progressive Generation Architecture}

The challenge of pseudo-query generation lies in simultaneously satisfying three orthogonal objectives, including
(i) style consistency, where queries must maintain consistent register with official examples,
(ii) capability coverage, where templates should cover as many distinct parsed capabilities as possible under the generation budget,
and (iii) parameter resolvability, where parameter slots must admit valid instantiation. 
Since a single LLM generation process is difficult to optimize these objectives simultaneously, we decouple the generation process into three progressive transformations:
\[
\Sigma_i = F_1(E_i),
\]
\[
T_i = F_2(\Sigma_i, C_i, P_i),
\]
\[
\widetilde{\mathcal{Q}}_i
=
F_3(T_i, P_i, \Sigma_i^{\mathrm{params}}),
\]
where \(F_1\), \(F_2\), and \(F_3\) correspond to the Style Mimicker,
Query Template Generator, and Param Filler, respectively. 

\subsubsection{Style Mimicker}

The Style Mimicker extracts a structured style representation from the
usage examples associated with skill \(s_i\):
\[
F_1: E_i \rightarrow \Sigma_i.
\]

The resulting representation
\[
\Sigma_i
=
\left(
\Sigma_i^{\mathrm{patterns}},
\Sigma_i^{\mathrm{words}},
\Sigma_i^{\mathrm{params}},
\Sigma_i^{\mathrm{examples}}
\right)
\]
contains four complementary components: syntactic patterns, domain-specific
vocabulary, parameter expressions, and raw example queries.

The syntactic-pattern component is defined as
\[
\Sigma_i^{\mathrm{patterns}}
=
\{(\pi_k,c_k)\}_{k=1}^{K},
\]
where \(\pi_k\) denotes a syntactic pattern and \(c_k\) records its frequency
among the skill examples.

The vocabulary component is defined as
\[
\Sigma_i^{\mathrm{words}}
=
\{\text{verbs},\text{domain terms},\ldots\},
\]
and stores part-of-speech-bucketed lexical items that characterize the
domain and expression style of the examples.

The parameter-expression component is represented as
\[
\Sigma_i^{\mathrm{params}}: P_i \rightarrow \mathcal{T},
\]
which maps each parameter to the natural-language expressions used to refer
to it in the examples, where \(\mathcal{T}\) denotes the space of textual
expressions.

Finally, \(\Sigma_i^{\mathrm{examples}}\) preserves the original example
queries as direct style references for subsequent query-template generation.

\subsubsection{Query Template Generation}

Given the style representation \(\Sigma_i\), the capability set \(C_i\), and
the parameter set \(P_i\), the Query Template Generator produces a set of
capability-aware query templates:
\[
F_2\colon
(\Sigma_i, C_i, P_i)
\rightarrow
T_i
=
\{
t_{i,1},
\ldots,
t_{i,K_i}
\}.
\]

Each template \(t_{i,j}\) is represented as a triple
\[
t_{i,j}
=
(
\text{text},
\text{cap},
\text{slots}
),
\]
where \(\text{text}\) denotes the natural-language template,
\(\text{cap}\) identifies the associated capability, and
\(\text{slots}\) contains parameter placeholders of the form
\(\{\text{param}\}\).

To ensure that all capabilities are represented in the generated template
set, we impose the following coverage constraint:
\[
\max_{T_i:\,|T_i|\le B}
\left|
\left\{
\operatorname{cap}(t)
\mid t\in T_i
\right\}
\right|,
\]
where \(B\) is the maximum number of templates generated for each skill.

The generated templates cover diverse sentence forms, including direct
imperatives, yes/no questions, wh-questions, and indirect requests, thereby
improving the linguistic diversity of the resulting pseudo-queries.

\subsubsection{Parameter Filling}

After query-template generation, the Param Filler instantiates the parameter
slots according to the parameter definitions and the expressions extracted
from skill examples:
\[
F_3\colon
(T_i, P_i, \Sigma_i^{\mathrm{params}})
\rightarrow
\widetilde{\mathcal{Q}}_i.
\]

For each parameter slot, candidate values are selected according to the
following priority order:
\[
\text{examples}
\rightarrow
\text{enum}
\rightarrow
\text{default}
\rightarrow
\text{type fallback}.
\]

For templates containing multiple parameters, candidate values are first combined through a Cartesian product and then capped at a preset number of combinations using priority-based sampling to prevent combinatorial growth.
The resulting pseudo-queries are then checked through static validation, including parameter-name matching, required-parameter coverage, type compatibility,
range constraints, and enumeration validity. The Param Filler adopts a
rule-based strategy to ensure that the generated pseudo-queries remain
consistent with the executable interface of the corresponding skill.
\subsection{Deployment of Generated Pseudo-Queries}

The generated pseudo-queries support three deployment scenarios. For
offline index augmentation, each pseudo-query $\widetilde{q}_{i,j}$ is
associated with its source skill $s_i$ and added to the retrieval index. Each pseudo-query is indexed separately under its source skill identifier, so retrieving it retrieves the corresponding skill.
For online query expansion, a seed skill retrieved from the original query
conditions the generation of query variants
\[
\mathcal{Q}_{\mathrm{exp}}
=
\{q,q_1,\ldots,q_n\},
\]
whose ranked lists are fused using Reciprocal Rank Fusion~\cite{cormack2009reciprocal}. For retriever
training, each pseudo-query is paired with its source skill to form
synthetic query--skill supervision:
\[
\mathcal{D}_{\mathrm{train}}
=
\{(\widetilde{q}_{i,j},s_i)\mid
\widetilde{q}_{i,j}\in\widetilde{\mathcal{Q}}_i\}.
\]

\begin{table*}[t]
\centering
\caption{Recall performance of Skill2Query in the offline mode.
Baseline results are shown outside parentheses, with offline gains
in percentage points shown in parentheses.}
\label{tab:offline_recall}

\resizebox{\textwidth}{!}{
\begin{tabular}{l|cc|cc|cc|cc}
\toprule
Method &
\multicolumn{2}{c|}{TheoremQA} &
\multicolumn{2}{c|}{LogicBench} &
\multicolumn{2}{c|}{ToolQA} &
\multicolumn{2}{c}{CHAMP}\\

&
R@1 & R@10 &
R@1 & R@10 &
R@1 & R@10 &
R@1 & R@10\\
\midrule

BM25 (full) 
& 60.37 (\(\pm\)0.00) & 83.94 (\(\pm\)0.00)
& 15.39 (+6.58) & 45.39 (+21.19)
& 46.15 (+2.17) & 81.61 (+2.45)
& 13.90 (-3.14) & 36.29 (-6.12)\\

BM25 (name)
& 16.06 (+13.12) & 29.45 (+22.22)
& 0.00 (+10.39) & 1.84 (+31.19)
& 5.52 (+26.51) & 24.27 (+37.27)
& 0.45 (+9.90) & 1.72 (+23.71)\\

BM25 (name+desc)
& 22.22 (+12.45) & 33.73 (+23.97)
& 3.16 (+6.31) & 13.95 (+18.42)
& 19.44 (+13.43) & 43.43 (+18.46)
& 3.81 (+6.28) & 7.44 (+18.79)\\

BGE-Large-v1.5 (full)
& 63.86 (+0.93) & 83.94 (+2.27)
& 7.89 (-1.97) & 31.45 (-8.82)
& 34.76 (-2.17) & 75.94 (-5.03)
& 10.74 (+0.10) & 43.36 (-2.20)\\

BGE-Large-v1.5 (name)
& 36.81 (+16.07) & 58.64 (+22.89)
& 2.76 (+1.85) & 11.58 (+8.03)
& 15.52 (+15.11) & 48.46 (+17.76)
& 5.31 (+5.79) & 18.24 (+16.36)\\

BGE-Large-v1.5 (name+desc)
& 50.60 (+2.55) & 75.37 (+7.36)
& 3.68 (+1.45) & 19.74 (-0.27)
& 25.87 (+4.41) & 64.69 (+2.16)
& 8.48 (+4.04) & 30.92 (+4.62)\\

SR-Emb-0.6B (full)
& 55.15 (+8.04) & 85.14 (+7.23)
& 5.39 (+4.61) & 30.79 (+5.92)
& 13.22 (+16.57) & 38.81 (+19.72)
& 16.59 (-2.05) & 49.68 (-2.74)\\

SR-Emb-0.6B (name)
& 52.07 (+15.80) & 82.33 (+9.90)
& 6.18 (+4.08) & 24.34 (+13.42)
& 19.72 (+12.38) & 49.65 (+12.52)
& 6.58 (+6.50) & 39.95 (+5.60)\\

SR-Emb-0.6B (name+desc)
& 54.89 (+8.83) & 87.42 (+5.62)
& 4.34 (+5.66) & 32.63 (+4.48)
& 7.76 (+21.75) & 25.59 (+32.10)
& 13.90 (+0.86) & 48.71 (-0.95)\\

SkillRouter (full)
& 82.73 (+1.07) & 91.40 (+3.11)
& 22.37 (-1.19) & 40.13 (+5.79)
& 35.80 (+11.54) & 49.44 (+9.44)
& 26.64 (+1.42) & 56.89 (+3.66)\\

\bottomrule
\end{tabular}
}
\end{table*}

\begin{table*}[t]
\centering
\caption{nDCG performance under offline index augmentation. Baseline scores
are shown outside parentheses, and absolute gains in percentage points are
shown in parentheses.}
\label{tab:offline_ndcg}

\resizebox{\textwidth}{!}{
\begin{tabular}{l|cc|cc|cc|cc}
\toprule
Method &
\multicolumn{2}{c|}{TheoremQA} &
\multicolumn{2}{c|}{LogicBench} &
\multicolumn{2}{c|}{ToolQA} &
\multicolumn{2}{c}{CHAMP}\\

&
N@1 & N@10 &
N@1 & N@10 &
N@1 & N@10 &
N@1 & N@10\\
\midrule

BM25 (full) 
& 60.37 ($\pm$0.00) & 71.85 ($\pm$0.00)
& 15.39 (+6.58) & 28.37 (+14.62)
& 46.15 (+2.17) & 64.92 (+3.43)
& 21.97 (-4.48) & 27.91 (-5.43)\\

BM25 (name) 
& 16.06 (+13.12) & 22.44 (+17.18)
& 0.00 (+10.39) & 0.88 (+20.15)
& 5.52 (+26.51) & 14.86 (+32.92)
& 0.90 (+16.59) & 1.25 (+18.14)\\

BM25 (name+desc)
& 22.22 (+12.45) & 27.33 (+18.16)
& 3.16 (+6.31) & 8.24 (+12.02)
& 19.44 (+13.43) & 32.05 (+15.12)
& 4.48 (+12.56) & 5.72 (+13.99)\\

BGE-Large-v1.5 (full)
& 63.86 (+0.93) & 73.90 (+1.36)
& 7.89 (-1.97) & 18.09 (-4.88)
& 34.76 (-2.17) & 57.29 (-5.11)
& 14.35 (+0.45) & 27.92 (-0.95)\\

BGE-Large-v1.5 (name)
& 36.81 (+16.07) & 47.06 (+19.89)
& 2.76 (+1.85) & 6.62 (+4.30)
& 15.52 (+15.11) & 33.01 (+15.65)
& 6.73 (+10.31) & 11.83 (+11.84)\\

BGE-Large-v1.5 (name+desc)
& 50.60 (+2.55) & 62.62 (+5.16)
& 3.68 (+1.45) & 10.65 (+0.49)
& 25.87 (+4.41) & 47.28 (+1.70)
& 10.76 (+7.63) & 19.49 (+5.25)\\

SR-Emb-0.6B (full)
& 55.15 (+8.04) & 69.57 (+8.49)
& 5.39 (+4.61) & 16.77 (+5.29)
& 13.22 (+16.57) & 25.12 (+19.04)
& 22.42 (-0.90) & 34.72 (-1.77)\\

SR-Emb-0.6B (name)
& 52.07 (+15.80) & 66.28 (+13.48)
& 6.18 (+4.08) & 13.87 (+8.60)
& 19.72 (+12.38) & 33.55 (+13.35)
& 8.52 (+10.76) & 22.99 (+6.53)\\

SR-Emb-0.6B (name+desc)
& 54.89 (+8.83) & 70.35 (+8.29)
& 4.34 (+5.66) & 16.31 (+5.89)
& 7.76 (+21.75) & 15.19 (+28.45)
& 18.83 (+3.14) & 32.48 (+1.22)\\

SkillRouter (full)
& 82.73 (+1.07) & 87.12 (+2.17)
& 22.37 (-1.19) & 31.64 (+1.01)
& 35.80 (+11.54) & 44.23 (+9.84)
& 38.12 (+1.79) & 46.37 (+2.08)\\

\bottomrule
\end{tabular}
}
\end{table*}

\section{Experiments and Results}

\subsection{Experimental Setup}

We evaluate Skill2Query on four datasets: TheoremQA~\cite{chen2023theoremqa}, LogicBench~\cite{parmar2024logicbench},
ToolQA~\cite{zhuang2023toolqa}, and
CHAMP~\cite{mao2024champ}. The evaluation split contains 3,160 instances, ranging from 223 in CHAMP to 1,430 in ToolQA. All queries are retrieved from a shared candidate pool of 26,262 skills. Both gold skill annotations and the candidate skill pool are drawn from SRA-Bench~\cite{su2026skill}. Each candidate skill is represented by its name, description, and body, which are concatenated to form the full skill text.

Experiments comprise two categories of comparisons. For generation quality, we
compare Skill2Query against Zero-shot LLM, Few-shot LLM~\cite{brown2020language}, and a SkillFlow-style
method~\cite{li2026skillflow}, using Exec-Pass, Func-Coverage, and Distinct-3.
For retrieval performance, we adopt BM25~\cite{robertson2009probabilistic},
BGE-Large-v1.5~\cite{xiao2024c}, SR-Emb-0.6B, and the two-stage
SkillRouter~\cite{zheng2026skillrouter}, comparing baseline, offline, and online
modes. We also fine-tune SkillRouter using pseudo-queries from different
methods to analyze training data quality impact.

Retrieval performance is evaluated using Recall@1, Recall@10, nDCG@1, and
nDCG@10~\cite{jarvelin2002cumulated}. For samples with multiple gold skills, Recall@\(k\) measures coverage
proportion, while nDCG@\(k\) uses binary relevance (gold \(=1\), non-gold \(=0\)).
All experiments run on \(4\times\) NVIDIA RTX 4090 24 GB GPUs. Detailed settings are provided in Appendices~A--D.

\subsection{RQ1: Pseudo-Query Generation Quality}

\begin{figure}[t]
    \centering
    \includegraphics[width=\columnwidth]{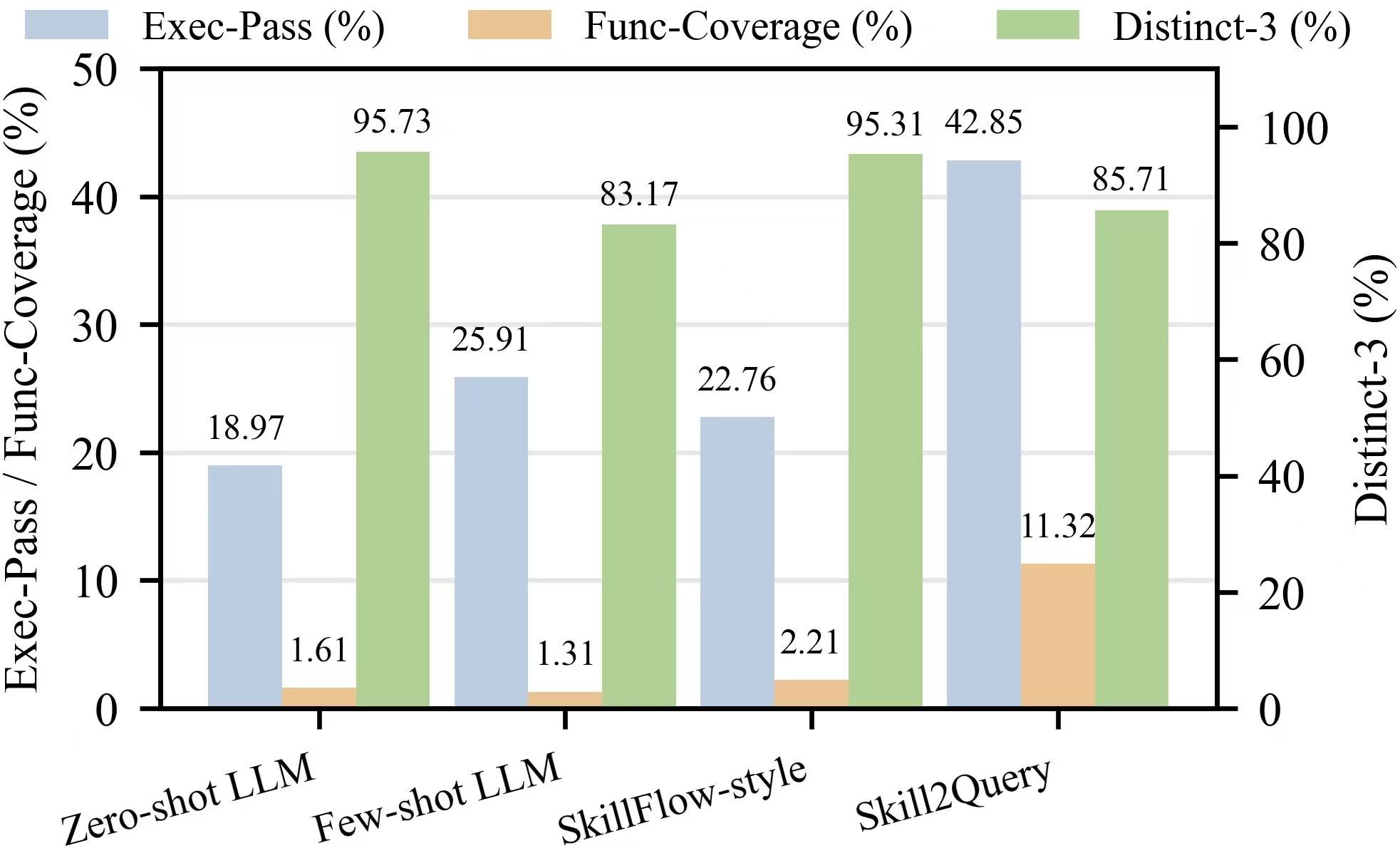}
    \caption{Pseudo-query generation quality comparison.}
    \Description{Comparison of different pseudo-query generation methods using Exec-Pass, Func-Coverage, and Distinct-3 metrics.}
    \label{fig:rq1}
\end{figure}

To analyze the impact of structured skill information on pseudo-query
generation quality, we design four representative approaches: Zero-shot LLM
(skill names + descriptions), Few-shot LLM (adds random examples),
SkillFlow-style (full text without structural parsing), and Skill2Query (parsed SKG). 
Detailed configurations are provided in Appendix~A. Figure~\ref{fig:rq1} presents the comparison across Exec-Pass, Func-Coverage, and Distinct-3.

Skill2Query achieves superior parameter validity and functional
coverage: Exec-Pass reaches \(42.85\%\) (improvements of \(16.94\) pp over
Few-shot, \(23.88\) pp over Zero-shot, and \(20.09\) pp over SkillFlow-style).
Func-Coverage reaches \(11.32\%\) (\(9.11\) pp over the best baseline). This
indicates SKG's structured information provides more explicit functional and
parametric constraints. In expression diversity, Skill2Query's Distinct-3 is \(85.71\%\), higher than
Few-shot (\(83.17\%\)) but lower than Zero-shot and SkillFlow-style, indicating
Skill2Query does not maximize surface-level variation but maintains diversity
under functional and parametric constraints. Ablation experiments are provided in Appendix~D.

\subsection{RQ2: Retrieval Performance Gains}

We evaluate Skill2Query with sparse retrieval (BM25), dense retrieval
(BGE-Large-v1.5 and SR-Emb-0.6B), and two-stage retrieval
(SkillRouter), under baseline, offline, and online settings. Tables~\ref{tab:offline_recall} and~\ref{tab:offline_ndcg} present the results of offline index augmentation. Offline index augmentation
yields positive gains across most datasets and retrievers. This effect is most
pronounced on ToolQA: BM25 name configuration raises R@1 from \(5.52\%\) to
\(32.03\%\), R@10 from \(24.27\%\) to \(61.54\%\), while N@1 and N@10
increase by \(26.51\) and \(32.92\) pp. Dense retrievers likewise benefit:
BGE name and SR-Emb-0.6B name+description achieve R@1 gains of \(15.11\) and
\(21.75\) pp on ToolQA. On two-stage SkillRouter, ToolQA R@1 improves from
\(35.80\%\) to \(47.34\%\), R@10 from \(49.44\%\) to \(58.88\%\), nDCG@10 increases by 9.84 pp. Improvements in R@10 often exceed R@1, indicating offline
augmentation more effectively brings gold skills into the candidate set. The
advantage is especially prominent when only skill names or brief descriptions
are used, confirming that structured pseudo-queries mitigate semantic
sparsity.

\begin{table}[t]
\centering
\setlength{\tabcolsep}{2pt}
\renewcommand{\arraystretch}{0.8}

\caption{Recall statistics of Skill2Query online query expansion. 
R@1 and R@10 are reported as percentages.}
\label{tab:online_recall}

\resizebox{\linewidth}{!}{
\begin{tabular}{l|cc|cc|cc|cc}
\toprule
Method &
\multicolumn{2}{c|}{TheoremQA} &
\multicolumn{2}{c|}{LogicBench} &
\multicolumn{2}{c|}{ToolQA} &
\multicolumn{2}{c}{CHAMP}\\

& R@1 & R@10 &
R@1 & R@10 &
R@1 & R@10 &
R@1 & R@10\\
\midrule

BM25 (baseline)
&60.37&83.94&15.39&45.39&46.15&81.61&13.90&36.29\\

BM25 (offline)
&60.37&83.94&21.97&66.58&48.32&84.06&10.76&30.17\\

BM25+S2Q online
&65.19&89.42&12.76&37.50&46.43&81.19&18.39&46.67\\

SkillRouter (baseline)
&82.73&91.40&22.37&40.13&35.80&49.44&26.64&56.89\\

SkillRouter (offline)
&83.80&94.51&21.18&45.92&47.34&58.88&28.06&60.55\\

SkillRouter+S2Q online
&81.24&93.31&18.42&36.32&32.03&47.83&26.23&60.18\\

SR-Emb-0.6B (baseline)
&55.15&85.14&5.39&30.79&13.22&38.81&16.59&49.68\\

SR-Emb-0.6B (offline)
&63.19&92.37&10.00&36.71&29.79&58.53&14.54&46.94\\

SR-Emb-0.6B+S2Q online
&57.56&89.83&4.34&19.61&11.12&35.38&17.45&53.38\\

\bottomrule
\end{tabular}
}
\end{table}

\begin{table}[t]
\centering
\setlength{\tabcolsep}{2pt}
\renewcommand{\arraystretch}{0.8}

\caption{nDCG statistics of Skill2Query online query expansion. N@1 and N@10 are reported as percentages.}
\label{tab:online_ndcg}

\resizebox{\linewidth}{!}{
\begin{tabular}{l|cc|cc|cc|cc}
\toprule
Method &
\multicolumn{2}{c|}{TheoremQA} &
\multicolumn{2}{c|}{LogicBench} &
\multicolumn{2}{c|}{ToolQA} &
\multicolumn{2}{c}{CHAMP}\\

& N@1 & N@10 &
N@1 & N@10 &
N@1 & N@10 &
N@1 & N@10\\
\midrule

BM25 (baseline)
&60.37&71.85&15.39&28.37&46.15&64.92&21.97&27.91\\

BM25 (offline)
&60.37&71.85&21.97&42.99&48.32&68.35&17.49&22.48\\

BM25+S2Q online
&65.19&77.32&12.76&23.83&46.43&64.83&28.70&35.60\\

SkillRouter (baseline)
&82.73&87.12&22.37&31.64&35.80&44.23&38.12&46.37\\

SkillRouter (offline)
&83.80&89.29&21.18&32.65&47.34&54.07&39.91&48.45\\

SkillRouter+S2Q online
&81.24&87.15&18.42&26.36&32.03&40.28&38.57&47.50\\

SR-Emb-0.6B (baseline)
&55.15&69.57&5.39&16.77&13.22&25.12&22.42&34.72\\

SR-Emb-0.6B (offline)
&63.19&78.06&10.00&22.06&29.79&44.16&21.52&32.95\\

SR-Emb-0.6B+S2Q online
&57.56&72.89&4.34&10.62&11.12&21.58&23.77&37.63\\

\bottomrule
\end{tabular}
}
\end{table}
For the online mode, Tables~\ref{tab:online_recall} and~\ref{tab:online_ndcg} show that Skill2Query dynamically generates
query variants conditioned on the current query and seed skill. Online mode
lifts BM25 R@1 on TheoremQA from \(60.37\%\) to \(65.19\%\), R@10 from
\(83.94\%\) to \(89.42\%\). On CHAMP, R@1 and N@10 improve from \(13.90\%\)
and \(27.91\%\) to \(18.39\%\) and \(35.60\%\), surpassing offline settings.
SR-Emb-0.6B online mode raises R@10 on CHAMP from \(49.68\%\) to \(53.38\%\).
The two modes are complementary: offline provides general, stable index
supplementation, and online tailors expressions to specific queries.

\subsection{RQ3: Quality of Pseudo-Query Training Data}

To assess Skill2Query-generated pseudo-queries as training data, we keep the retrieval architecture, training objectives, and hyperparameters fixed while varying the pseudo-query generation method, namely Few-shot LLM, SkillFlow-style, and Skill2Query. 
All methods construct query--skill training pairs and use identical fine-tuning procedures to isolate training data quality
effects (Appendix~C).

\begin{table}[t]
\centering
\setlength{\tabcolsep}{3pt}
\renewcommand{\arraystretch}{0.85}

\caption{Recall results across four benchmarks (training data comparison).}
\label{tab:training_recall}

\resizebox{\linewidth}{!}{
\begin{tabular}{l|cc|cc|cc|cc}
\toprule
Method &
\multicolumn{2}{c|}{TheoremQA} &
\multicolumn{2}{c|}{LogicBench} &
\multicolumn{2}{c|}{ToolQA} &
\multicolumn{2}{c}{CHAMP}\\

& R@1 & R@10 &
R@1 & R@10 &
R@1 & R@10 &
R@1 & R@10\\
\midrule

Few-shot
&82.06&96.25&18.68&38.55&40.91&47.20&30.04&57.06\\

SkillFlow-style
&83.40&96.52&19.08&42.11&41.40&45.87&28.29&50.34\\

Skill2Query
&85.27&95.58&21.97&49.34&45.03&52.31&31.73&62.22\\

\bottomrule
\end{tabular}
}
\end{table}

\begin{table}[t]
\centering
\setlength{\tabcolsep}{3pt}
\renewcommand{\arraystretch}{0.85}

\caption{nDCG results across four benchmarks (training data comparison).}
\label{tab:training_ndcg}
\resizebox{\linewidth}{!}{
\begin{tabular}{l|cc|cc|cc|cc}
\toprule
Method &
\multicolumn{2}{c|}{TheoremQA} &
\multicolumn{2}{c|}{LogicBench} &
\multicolumn{2}{c|}{ToolQA} &
\multicolumn{2}{c}{CHAMP}\\

& N@1 & N@10 &
N@1 & N@10 &
N@1 & N@10 &
N@1 & N@10\\
\midrule

Few-shot
&82.06&89.72&18.68&28.48&40.91&44.83&45.74&48.32\\

SkillFlow-style
&83.40&90.41&19.08&30.39&41.40&44.22&42.60&43.77\\

Skill2Query
&85.27&90.82&21.97&34.84&45.03&49.34&47.98&52.22\\

\bottomrule
\end{tabular}
}
\end{table}

\begin{figure}[t]
    \centering
    \includegraphics[width=\columnwidth]
     {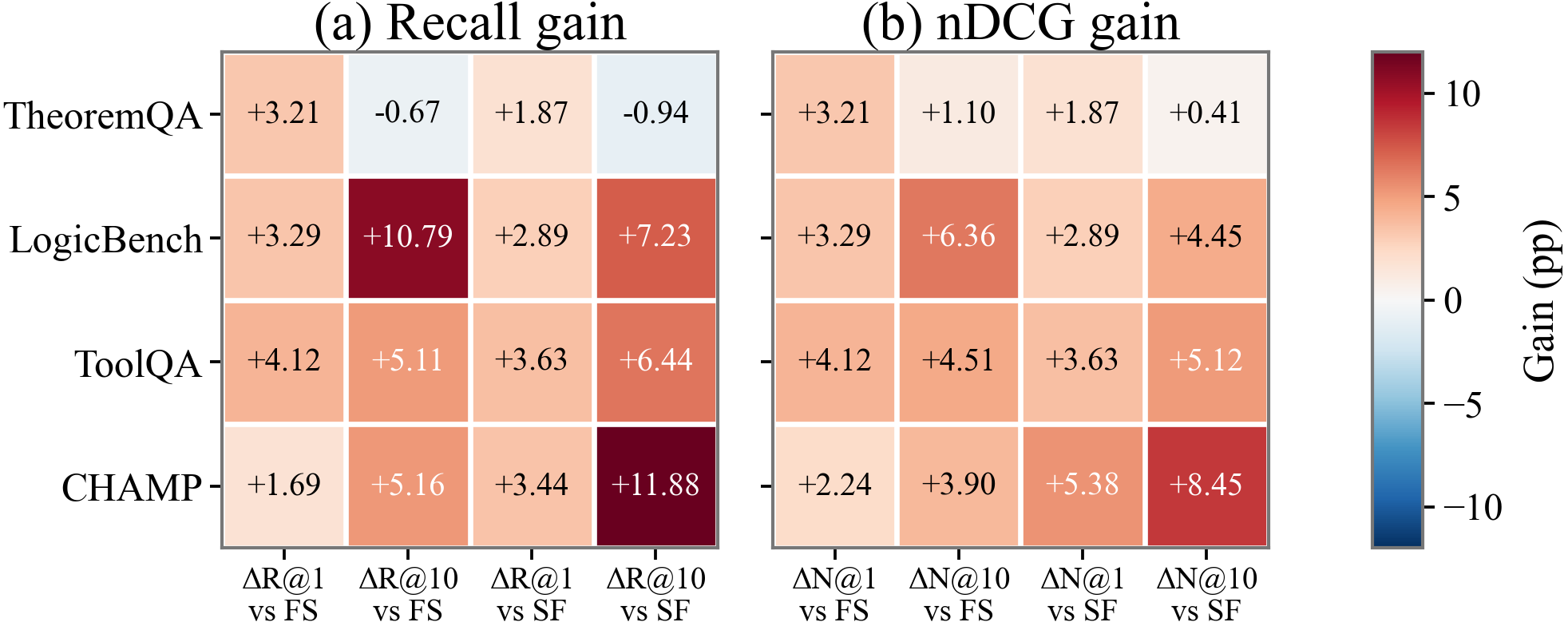}
    \caption{Performance gains of Skill2Query across datasets and retrieval metrics.}
    \Description{The heatmap illustrates the retrieval performance improvements of Skill2Query-generated training data compared with Few-shot LLM and SkillFlow-style methods across different datasets and evaluation metrics.}
    \label{fig:rq3}
\end{figure}

Tables~\ref{tab:training_recall} and~\ref{tab:training_ndcg} compare retrieval performance with the same SkillRouter 1.2B
two-stage architecture. Figure~\ref{fig:rq3} further visualizes the retrieval performance differences through a heatmap, providing an intuitive comparison of training data quality across different datasets.
Skill2Query training data achieves the best R@1 and
N@1 across all four datasets, with more pronounced advantages on
LogicBench, ToolQA, and CHAMP. LogicBench: R@10 \(49.34\%\) (\(+10.79\) pp
over Few-shot, \(+7.23\) pp over SkillFlow-style), N@10 \(34.84\%\).
CHAMP: R@10 \(62.22\%\) (\(+5.16\) pp, \(+11.88\) pp), N@10 \(52.22\%\).
On TheoremQA: R@1 \(85.27\%\), N@10 \(90.82\%\). These results validate
structured pseudo-queries as retrieval training data.

\subsection{Case Study: End-to-End Task Performance Validation}

\begin{table}[t]
\centering
\caption{End-to-end task success rates across LLM backends under different skill-retrieval configurations (\%). FT denotes fine-tuning with
Skill2Query-generated pseudo-queries.}
\label{tab:end2end}

\resizebox{\linewidth}{!}{
\begin{tabular}{l|l|c|cc|c}
\toprule
Agent &
Retriever &
No Skill &
Baseline &
Offline &
Oracle \\
\midrule

\multirow{3}{*}{GPT-5.4}
& BM25
& \multirow{3}{*}{80.05}
& 82.73 & 83.80
& \multirow{3}{*}{84.07} \\

& SR-Emb-0.6B
&
& 80.32 & 82.58
& \\

& SkillRouter (FT)
&
& 83.80 & 83.53
& \\
\midrule

\multirow{3}{*}{DeepSeek-V4-Flash}
& BM25
& \multirow{3}{*}{73.76}
& 76.31 & 76.17
& \multirow{3}{*}{85.69} \\

& SR-Emb-0.6B
&
& 73.36 & 77.11
& \\

& SkillRouter (FT)
&
& 78.18 & 82.33
& \\
\midrule

\multirow{3}{*}{Qwen3.6-Plus}
& BM25
& \multirow{3}{*}{79.79}
& 82.33 & 83.00
& \multirow{3}{*}{83.27} \\

& SR-Emb-0.6B
&
& 80.86 & 81.93
& \\

& SkillRouter (FT)
&
& 82.73 & 83.13
& \\
\bottomrule
\end{tabular}
}
\end{table}

To assess whether retrieval quality translates into downstream gains, we conduct an end-to-end study on 747 single-gold TheoremQA instances. For each instance, the full top-1 skill is injected into the prompt for one-step answering, with final accuracy as the metric and Oracle gold-skill injection as the upper bound. 

Table~\ref{tab:end2end} presents end-to-end task success rates. In the offline setting, the SkillRouter retriever fine-tuned in our study improves downstream accuracy over the no-skill baseline by \(3.48\) pp for GPT-5.4, \(8.57\) pp for DeepSeek-V4-Flash, and \(3.34\) pp for Qwen3.6-Plus. Overall, it delivers the strongest non-oracle performance in this setting. On GPT-5.4 and Qwen3.6-Plus, Skill2Query-based retrieval achieves
\(83.53\%\) and \(83.13\%\), differing from Oracle by only \(0.54\) and
\(0.14\) pp. For DeepSeek-V4-Flash, success rate rises from \(73.76\%\) to
\(82.33\%\) (\(+8.57\) pp). When the base model's reasoning capacity is
limited, high-quality skill retrieval provides more effective capability
supplementation.

\section{Discussion}

The core design of Skill2Query, which parses first and then generates, rests on a
fundamental observation: the representation gap in skill retrieval is not
merely semantic, but perspectival. Developers write SKILL.md for
functional completeness, emphasizing ``what I can provide,'' whereas users
express task goals, emphasizing ``what I want to accomplish.'' This creates
systematic differences in vocabulary, abstraction, and focus. SKG provides a
shared reference frame: Capability nodes extract functional atoms, Example
nodes capture user expression patterns, and Parameter nodes define the
executable interface between them.

The three-stage progressive generation is effective not because it ``does
more'' than a single LLM call, but because it allocates different dimensions of
the generation process to the information sources that genuinely contain the
relevant information: style comes from examples, template structure from
capability lists, and parameter values from schema definitions. 
Few-shot LLM mixes all information sources into a single prompt. The gap between its \(25.91\%\) Exec-Pass and Skill2Query's \(42.85\%\) measures precisely the degree to which information-source confusion degrades generation quality.

The complementary relationship between offline augmentation and online
expansion reveals a more general trade-off~\cite{weller2024generative}. Offline augmentation is
pre-computed semantic bridging, completing user expression generation once
during index construction, trading storage cost for zero query latency. Online
expansion is dynamic semantic bridging, adjusting the bridging strategy in
real time, trading computation latency for contextual adaptability. The BM25
full offline configuration yielded no gain on TheoremQA, while online mode
raised R@1 from \(60.37\%\) to \(65.19\%\), illustrating that when the original
index already contains rich semantic signals, the marginal value of static
supplementation decreases. This suggests the optimal switch point depends on
the semantic richness of the skill documents themselves.

The RQ3 results point to a conclusion with methodological significance:
training data quality depends not only on query--skill relevance, but also on
the granularity of that relevance. Queries from Few-shot LLM and SkillFlow-style
are correlated with overall skill semantics but cannot be traced to specific
Capabilities. Each Skill2Query pseudo-query is bound to a concrete functional
point via templates and to a valid parameter space via parameter slots. This
fine-grained correspondence forces the retrieval model to learn more
discriminative matching patterns, identifying ``which function this query
corresponds to and what parameters it requires'' rather than just ``whether
this query relates to this skill.'' The R@10 gains of \(10.79\) pp on LogicBench
and \(11.88\) pp on CHAMP, far exceeding TheoremQA's modest improvement, are
consistent with the skill complexity ranking of the datasets.

The end-to-end case study provides preliminary evidence that retrieval improvements can translate into task-level gains under the evaluated agent configurations.
Moreover, a complementary relationship exists between
base model capability and external skill retrieval: when the model's own
ability is limited, accurate skill retrieval provides more pronounced
capability supplementation. On stronger models, retrieval approaching Oracle
reduces interference from incorrect skills. Skill retrieval is therefore not an
isolated intermediate module. Its quality directly affects whether an agent
can convert external skills into task-completion capability~\cite{molfetta2025ports}.

This work has several limitations. The parsing quality of SKG is constrained by
the standardization of the original skill documentation. When documents lack
parameter declarations or usage examples, generation quality degrades to near
the Few-shot baseline. This reveals a deeper issue: the efficacy ceiling of
Skill2Query is determined not by the generation algorithm, but by the degree of
documentation standardization in the skill ecosystem. In an open skill
community with uneven documentation quality, the variance in generation
quality may exceed the mean difference between different generation methods.
Furthermore, applicability to larger-scale skill repositories, different agent
frameworks, and real user queries requires further investigation.
\section{Conclusion}

This paper presents Skill2Query, a structured pseudo-query generation method for
skill retrieval, designed to alleviate the scarcity of high-quality query--skill
training data. The method leverages a skill knowledge graph to organize
information about skill functions, parameters, and examples, and through query
template generation, parameter filling, and validity checking, converts
developer-side skill definitions into pseudo-queries that approximate user
expressions. Experimental results demonstrate consistent improvements across
multiple dimensions, e.g., pseudo-query quality, offline index augmentation, online
query expansion, and retrieval model training, and further improve the
end-to-end success rate of agents on real tasks, validating the significant
value of high-quality pseudo-queries for skill retrieval and skill invocation.

Future work will focus on user- and context-aware query generation mechanisms,
incorporating information such as user history and dialogue context into the
pseudo-query generation process to improve personalization and generalization
capability. Additionally, we will further investigate the co-optimization of
offline index augmentation and online query expansion, multi-skill joint
retrieval, and cross-platform skill transfer, and leverage real interaction
data to continuously refine skill indexing and training data, providing a more
efficient and scalable infrastructure for large-scale agent skill retrieval
systems.



\bibliographystyle{ACM-Reference-Format}
\bibliography{references}

@article{patil2024gorilla,
  title={Gorilla: Large language model connected with massive apis},
  author={Patil, Shishir G and Zhang, Tianjun and Wang, Xin and Gonzalez, Joseph E},
  journal={Advances in Neural Information Processing Systems},
  volume={37},
  pages={126544--126565},
  year={2024}
}

@inproceedings{qin2024toolllm,
  title={Toolllm: Facilitating large language models to master 16000+ real-world apis},
  author={Qin, Yujia and Liang, Shihao and Ye, Yining and Zhu, Kunlun and Yan, Lan and Lu, Yaxi and Lin, Yankai and Cong, Xin and Tang, Xiangru and Qian, Bill and others},
  booktitle={International Conference on Learning Representations},
  volume={2024},
  pages={9695--9717},
  year={2024}
}

@inproceedings{reimers2019sentence,
  title={Sentence-bert: Sentence embeddings using siamese bert-networks},
  author={Reimers, Nils and Gurevych, Iryna},
  booktitle={Proceedings of the 2019 conference on empirical methods in natural language processing and the 9th international joint conference on natural language processing (EMNLP-IJCNLP)},
  pages={3982--3992},
  year={2019}
}

@article{nogueira2019document,
  title={Document expansion by query prediction},
  author={Nogueira, Rodrigo and Yang, Wei and Lin, Jimmy and Cho, Kyunghyun},
  journal={arXiv preprint arXiv:1904.08375},
  year={2019}
}

@article{bonifacio2022inpars,
  title={Inpars: Data augmentation for information retrieval using large language models},
  author={Bonifacio, Luiz and Abonizio, Hugo and Fadaee, Marzieh and Nogueira, Rodrigo},
  journal={arXiv preprint arXiv:2202.05144},
  year={2022}
}

@inproceedings{li2023api,
  title={Api-bank: A comprehensive benchmark for tool-augmented llms},
  author={Li, Minghao and Zhao, Yingxiu and Yu, Bowen and Song, Feifan and Li, Hangyu and Yu, Haiyang and Li, Zhoujun and Huang, Fei and Li, Yongbin},
  booktitle={Proceedings of the 2023 conference on empirical methods in natural language processing},
  pages={3102--3116},
  year={2023}
}

@inproceedings{gao2023precise,
  title={Precise zero-shot dense retrieval without relevance labels},
  author={Gao, Luyu and Ma, Xueguang and Lin, Jimmy and Callan, Jamie},
  booktitle={Proceedings of the 61st Annual Meeting of the Association for Computational Linguistics (Volume 1: Long Papers)},
  pages={1762--1777},
  year={2023}
}

@inproceedings{wang2023query2doc,
  title={Query2doc: Query expansion with large language models},
  author={Wang, Liang and Yang, Nan and Wei, Furu},
  booktitle={Proceedings of the 2023 Conference on Empirical Methods in Natural Language Processing},
  pages={9414--9423},
  year={2023}
}

@inproceedings{wu2024seal,
  title={Seal-tools: Self-instruct tool learning dataset for agent tuning and detailed benchmark},
  author={Wu, Mengsong and Zhu, Tong and Han, Han and Tan, Chuanyuan and Zhang, Xiang and Chen, Wenliang},
  booktitle={CCF International Conference on Natural Language Processing and Chinese Computing},
  pages={372--384},
  year={2024},
  organization={Springer}
}

@article{zheng2025can,
  title={Can Synthetic Query Rewrites Capture User Intent Better than Humans in Retrieval-Augmented Generation?},
  author={Zheng, JiaYing and Zhang, HaiNan and Pang, Liang and Tong, YongXin and Zheng, ZhiMing},
  journal={arXiv preprint arXiv:2509.22325},
  year={2025}
}

@inproceedings{huang2023let,
  title={Let's chat to find the apis: Connecting human, llm and knowledge graph through ai chain},
  author={Huang, Qing and Wan, Zhenyu and Xing, Zhenchang and Wang, Changjing and Chen, Jieshan and Xu, Xiwei and Lu, Qinghua},
  booktitle={2023 38th IEEE/ACM International Conference on Automated Software Engineering (ASE)},
  pages={471--483},
  year={2023},
  organization={IEEE}
}

@misc{openapi2021specification,
  author       = {{OpenAPI Initiative}},
  title        = {{OpenAPI Specification, Version 3.1.0}},
  year         = {2021},
  howpublished = {\url{https://spec.openapis.org/oas/v3.1.0}},
  note         = {Accessed: 2026-07-25}
}

@article{ni2025toolfactory,
  title={Toolfactory: Automating tool generation by leveraging llm to understand rest api documentations},
  author={Ni, Xinyi and Wang, Qiuyang and Zhang, Yukun and Hong, Pengyu},
  journal={arXiv preprint arXiv:2501.16945},
  year={2025}
}

@inproceedings{li2026skillflow,
  title={Skillflow: Scalable and efficient agent skill retrieval system},
  author={Li, Fangzhou and Tagkopoulos, Pagkratios and Tagkopoulos, Ilias},
  booktitle={Deep Learning for Code: Towards Human-Centered Coding Agents},
  year={2026}
}

@book{robertson2009probabilistic,
  title={The probabilistic relevance framework: BM25 and beyond},
  author={Robertson, Stephen and Zaragoza, Hugo},
  volume={4},
  year={2009},
  publisher={Now Publishers Inc}
}

@inproceedings{xiao2024c,
  title={C-pack: Packed resources for general chinese embeddings},
  author={Xiao, Shitao and Liu, Zheng and Zhang, Peitian and Muennighoff, Niklas and Lian, Defu and Nie, Jian-Yun},
  booktitle={Proceedings of the 47th international ACM SIGIR conference on research and development in information retrieval},
  pages={641--649},
  year={2024}
}

@article{cho2026skillret,
  title={SkillRet: A large-scale benchmark for skill retrieval in LLM agents},
  author={Cho, Hongcheol and Kang, Ryangkyung and Kim, Youngeun},
  journal={arXiv preprint arXiv:2605.05726},
  year={2026}
}

@inproceedings{wang2022gpl,
  title={GPL: Generative pseudo labeling for unsupervised domain adaptation of dense retrieval},
  author={Wang, Kexin and Thakur, Nandan and Reimers, Nils and Gurevych, Iryna},
  booktitle={Proceedings of the 2022 conference of the North American chapter of the association for computational linguistics: human language technologies},
  pages={2345--2360},
  year={2022}
}

@inproceedings{dai2022promptagator,
  title={Promptagator: Few-shot dense retrieval from 8 examples},
  author={Dai, Zhuyun and Zhao, Vincent Y and Ma, Ji and Luan, Yi and Ni, Jianmo and Lu, Jing and Bakalov, Anton and Guu, Kelvin and Hall, Keith and Chang, Ming-Wei},
  booktitle={The Eleventh International Conference on Learning Representations},
  year={2022}
}

@inproceedings{chen2023theoremqa,
  title={Theoremqa: A theorem-driven question answering dataset},
  author={Chen, Wenhu and Yin, Ming and Ku, Max and Lu, Pan and Wan, Yixin and Ma, Xueguang and Xu, Jianyu and Wang, Xinyi and Xia, Tony},
  booktitle={Proceedings of the 2023 Conference on Empirical Methods in Natural Language Processing},
  pages={7889--7901},
  year={2023}
}

@inproceedings{kachuee2025improving,
  title={Improving tool retrieval by leveraging large language models for query generation},
  author={Kachuee, Mohammad and Ahuja, Sarthak and Kumar, Vaibhav and Xu, Puyang and Liu, Xiaohu},
  booktitle={Proceedings of the 31st International Conference on Computational Linguistics: Industry Track},
  pages={29--38},
  year={2025}
}

@inproceedings{parmar2024logicbench,
  title={Logicbench: Towards systematic evaluation of logical reasoning ability of large language models},
  author={Parmar, Mihir and Patel, Nisarg and Varshney, Neeraj and Nakamura, Mutsumi and Luo, Man and Mashetty, Santosh and Mitra, Arindam and Baral, Chitta},
  booktitle={Proceedings of the 62nd Annual Meeting of the Association for Computational Linguistics (Volume 1: Long Papers)},
  pages={13679--13707},
  year={2024}
}

@article{zhuang2023toolqa,
  title={Toolqa: A dataset for llm question answering with external tools},
  author={Zhuang, Yuchen and Yu, Yue and Wang, Kuan and Sun, Haotian and Zhang, Chao},
  journal={Advances in Neural Information Processing Systems},
  volume={36},
  pages={50117--50143},
  year={2023}
}

@inproceedings{mao2024champ,
  title={CHAMP: A Competition-Level Dataset for Fine-Grained Analyses of LLMs' Mathematical Reasoning Capabilities},
  author={Mao, Yujun and Kim, Yoon and Zhou, Yilun},
  booktitle={Findings of the Association for Computational Linguistics: ACL 2024},
  year={2024}
}

@article{liu2026graph,
  title={Graph-of-Skills: Dependency-Aware Structural Retrieval for Massive Agent Skills},
  author={Liu, Dawei and Li, Zongxia and Du, Hongyang and Wu, Xiyang and Gui, Shihang and Kuang, Yongbei and Sun, Lichao},
  journal={arXiv preprint arXiv:2604.05333},
  year={2026}
}

@article{zheng2026skillrouter,
  title={Skillrouter: Retrieve-and-rerank skill selection for llm agents at scale},
  author={Zheng, YanZhao and Zhang, ZhenTao and Ma, Chao and Yu, YuanQiang and Zhu, JiHuan and Dong, Baohua and Zhu, Hangcheng},
  journal={arXiv preprint arXiv:2603.22455},
  volume={4},
  year={2026}
}

@article{su2026skill,
  title={Skill retrieval augmentation for agentic ai},
  author={Su, Weihang and Long, Jianming and Ai, Qingyao and He, Qiaozhi and Tang, Yichen and Wang, Changyue and Tu, Yiteng and Wang, Yingbo and Liu, Yiqun},
  journal={arXiv preprint arXiv:2604.24594},
  year={2026}
}

@article{zhou2026externalization,
  title={Externalization in llm agents: A unified review of memory, skills, protocols and harness engineering},
  author={Zhou, Chenyu and Chai, Huacan and Chen, Wenteng and Guo, Zihan and Shan, Rong and Song, Yuanyi and Xu, Tianyi and Yang, Yingxuan and Yu, Aofan and Zhang, Weiming and others},
  journal={arXiv preprint arXiv:2604.08224},
  year={2026}
}

@article{chen2026skilljuror,
  title={SkillJuror: Measuring How Agent Skill Organization Changes Runtime Behavior},
  author={Chen, Zhiyu and Guo, Zihan and Huang, Bo and Lu, Bingwei and Lin, Jianghao and Zhou, Yuanjian and Zhang, Weinan},
  journal={arXiv preprint arXiv:2606.11543},
  year={2026}
}

@article{yu2026latentskill,
  title={Latentskill: From in-context textual skills to in-weight latent skills for llm agents},
  author={Yu, Aofan and Zhou, Chenyu and Xu, Tianyi and Guo, Zihan and Shan, Rong and Fu, Zhihui and Wang, Jun and Liu, Weiwen and Yu, Yong and Zhang, Weinan and others},
  journal={arXiv preprint arXiv:2606.06087},
  year={2026}
}

@article{guo2026skillprobe,
  title={Skillprobe: Security auditing for emerging agent skill marketplaces via multi-agent collaboration},
  author={Guo, Zihan and Chen, Zhiyu and Nie, Xiaohang and Lin, Jianghao and Zhou, Yuanjian and Zhang, Weinan},
  journal={arXiv preprint arXiv:2603.21019},
  year={2026}
}

@article{pan2026skillmas,
  title={Skillmas: Skill co-evolution with llm-based multi-agent system},
  author={Pan, Shuai and Liu, Yixiang and Gao, Jiaye and Gao, Te and Liu, Weiwen and Lin, Jianghao and Fu, Zhihui and Wang, Jun and Zhang, Weinan and Yu, Yong},
  journal={arXiv preprint arXiv:2605.09341},
  year={2026}
}

@article{wang2026skills,
  title={Skills on the Fly: Test-Time Adaptive Skill Synthesis for LLM Agents},
  author={Wang, Jingxing and Zhou, Chenyu and Fu, Zhihui and Wang, Jun and Liu, Weiwen and Zhang, Weinan and Lin, Jianghao},
  journal={arXiv preprint arXiv:2605.16986},
  year={2026}
}

@article{chen2408re,
  title={Re-invoke: tool invocation rewriting for zero-shot tool retrieval (2024)},
  author={Chen, Yanfei and Yoon, Jinsung and Sachan, Devendra Singh and Wang, Qingze and Cohen-Addad, Vincent and Bateni, Mohammadhossein and Lee, Chen-Yu and Pfister, Tomas},
  journal={arXiv preprint arXiv:2408.01875}
}

@inproceedings{cormack2009reciprocal,
  title={Reciprocal rank fusion outperforms condorcet and individual rank learning methods},
  author={Cormack, Gordon V and Clarke, Charles LA and Buettcher, Stefan},
  booktitle={Proceedings of the 32nd international ACM SIGIR conference on Research and development in information retrieval},
  pages={758--759},
  year={2009}
}

@inproceedings{li2016diversity,
  title={A diversity-promoting objective function for neural conversation models},
  author={Li, Jiwei and Galley, Michel and Brockett, Chris and Gao, Jianfeng and Dolan, William B},
  booktitle={Proceedings of the 2016 conference of the North American chapter of the association for computational linguistics: human language technologies},
  pages={110--119},
  year={2016}
}

@article{oord2018representation,
  title={Representation learning with contrastive predictive coding},
  author={Oord, Aaron van den and Li, Yazhe and Vinyals, Oriol},
  journal={arXiv preprint arXiv:1807.03748},
  year={2018}
}

@article{lin2025masstool,
  title={MassTool: A Multi-Task Search-Based Tool Retrieval Framework for Large Language Models},
  author={Lin, Jianghao and Wang, Xinyuan and Dai, Xinyi and Zhu, Menghui and Chen, Bo and Tang, Ruiming and Yu, Yong and Zhang, Weinan},
  journal={arXiv preprint arXiv:2507.00487},
  year={2025}
}

@inproceedings{shi2025retrieval,
  title={Retrieval models aren’t tool-savvy: Benchmarking tool retrieval for large language models},
  author={Shi, Zhengliang and Wang, Yuhan and Yan, Lingyong and Ren, Pengjie and Wang, Shuaiqiang and Yin, Dawei and Ren, Zhaochun},
  booktitle={Findings of the Association for Computational Linguistics: ACL 2025},
  pages={24497--24524},
  year={2025}
}

@article{jarvelin2002cumulated,
  title={Cumulated gain-based evaluation of IR techniques},
  author={J{\"a}rvelin, Kalervo and Kek{\"a}l{\"a}inen, Jaana},
  journal={ACM Transactions on Information Systems (TOIS)},
  volume={20},
  number={4},
  pages={422--446},
  year={2002},
  publisher={ACM New York, NY, USA}
}

@inproceedings{weller2024generative,
  title={When do generative query and document expansions fail? a comprehensive study across methods, retrievers, and datasets},
  author={Weller, Orion and Lo, Kyle and Wadden, David and Lawrie, Dawn and Van Durme, Benjamin and Cohan, Arman and Soldaini, Luca},
  booktitle={Findings of the Association for Computational Linguistics: EACL 2024},
  pages={1987--2003},
  year={2024}
}

@inproceedings{wang2023self,
  title={Self-instruct: Aligning language models with self-generated instructions},
  author={Wang, Yizhong and Kordi, Yeganeh and Mishra, Swaroop and Liu, Alisa and Smith, Noah A and Khashabi, Daniel and Hajishirzi, Hannaneh},
  booktitle={Proceedings of the 61st annual meeting of the association for computational linguistics (volume 1: long papers)},
  pages={13484--13508},
  year={2023}
}

@inproceedings{molfetta2025ports,
  title={PORTS: Preference-Optimized Retrievers for Tool Selection with Large Language Models},
  author={Molfetta, Lorenzo and Frisoni, Giacomo and Monaldini, Nicol{\`o} and Moro, Gianluca},
  booktitle={Proceedings of the 2025 Conference on Empirical Methods in Natural Language Processing},
  pages={10018--10041},
  year={2025}
}

@inproceedings{jiang2023noisy,
  title={Noisy self-training with synthetic queries for dense retrieval},
  author={Jiang, Fan and Drummond, Tom and Cohn, Trevor},
  booktitle={Findings of the Association for Computational Linguistics: EMNLP 2023},
  pages={11991--12008},
  year={2023}
}

@article{brown2020language,
  title={Language models are few-shot learners},
  author={Brown, Tom and Mann, Benjamin and Ryder, Nick and Subbiah, Melanie and Kaplan, Jared D and Dhariwal, Prafulla and Neelakantan, Arvind and Shyam, Pranav and Sastry, Girish and Askell, Amanda and others},
  journal={Advances in neural information processing systems},
  volume={33},
  pages={1877--1901},
  year={2020}
}

\appendix

\section{Pseudo-Query Generation Experiment Details}

This appendix supplements the implementation details of the pseudo-query generation pipeline evaluated in RQ1. All experiments are conducted on the same Skill Pool constructed in this work. 
All methods share the same skill set, preprocessing procedure, output schema, 
template validation module, pseudo-query instantiation module, deduplication procedure, 
and evaluation metrics. They differ only in the input information provided to the query 
template generator, ranging from plain metadata to full unstructured documents and structured SKGs. During offline corpus construction, we generate at most five parameterized query templates for each skill. All LLM generation steps use GPT-4o-mini with temperature 0.3.

\subsection{Baseline Methods and Input Construction}
\begin{table}[t]
\centering
\caption{Input information configurations of different pseudo-query generation methods.}
\label{tab:rq1_input_config}
\small
\begin{tabular}{p{0.18\columnwidth}p{0.25\columnwidth}p{0.45\columnwidth}}
\toprule
Method & Input Information & Description \\
\midrule

Zero-shot LLM 
& name + description 
& Uses only concise skill metadata without structured information or query examples. \\

Few-shot LLM 
& name + description + query examples 
& Adds three randomly sampled query examples from the same skill document, without explicit capability or parameter schema information. \\

SkillFlow-style 
& name + description + body 
& Uses the complete skill document as unstructured text without explicitly decomposing capabilities, parameter schemas, or examples. \\

Skill2Query 
& parsed SKG 
& Uses structured capability, parameter schema, and example information extracted from SKG. \\

\bottomrule
\end{tabular}
\end{table}
To analyze the impact of structured information on pseudo-query generation quality, we compare several generation methods, as shown in Table~\ref{tab:rq1_input_config}. All methods adopt the same generation template, parsing procedure, parameter filling strategy, post-processing pipeline, and metric computation process. The only difference lies in the input information and its organization format. Zero-shot LLM uses only the skill name and description as input. Few-shot LLM augments the skill metadata with up to three randomly sampled usage examples extracted from the same skill document. SkillFlow-style feeds the complete skill document as unstructured text.
Skill2Query uses the parsed SKG, which explicitly organizes capability information, parameter schemas, and usage examples to guide query template generation and parameter instantiation.
\subsection{Query Template Generation Prompt}

To ensure fair comparison among different generation methods, all approaches reuse the same Query Template Generator and adopt identical system prompts, output formats, and post-processing procedures. The shared system prompt used for query template generation is shown in Table~\ref{tab:query_generation_prompt}.

\begin{table}[t]
\centering
\caption{System prompt used for query template generation.}
\label{tab:query_generation_prompt}
\begin{minipage}{0.95\columnwidth}
\small
\hrule
\vspace{4pt}

\textbf{System Prompt}

\vspace{4pt}

You generate diverse English natural-language query templates for a skill-based retrieval system.

\vspace{4pt}

A template is a query a real user would type, with parameter values replaced by \{param\_name\} placeholders.

\vspace{4pt}

Rules:

1. Cover every capability at least once and distribute templates across capabilities.

2. Vary sentence forms, including direct commands, yes/no questions, wh-questions, and indirect requests.

3. Keep queries concise and realistic. Do not include explanations or meta-commentary.

4. Use only the exact \{param\_name\} placeholders listed in the input. Do not abbreviate, rewrite, or invent parameter names.

5. Every template must contain all required parameters. Optional parameters may be omitted.

6. All templates must be written in English.

\vspace{4pt}
\hrule
\end{minipage}
\end{table}

Under these constraints, the generator produces concise, realistic, and structurally valid query templates while following the provided capability and parameter information.

\subsection{Parameter Filling and Pseudo-Query Instantiation}

The Query Template Generator first produces natural-language query templates with parameter placeholders. For example:

\textit{How many ways can I seat \{n\} people at \{k\} identical round tables?}
where \{n\} and \{k\} represent parameter slots to be instantiated. All generation methods reuse the same QueryGenerationPipeline and follow three sequential stages: query template generation, parameter slot filling, and pseudo-query expansion. For Zero-shot LLM, parameter slots are mainly inferred by the generation model based on the skill name and description. Few-shot LLM further leverages a small number of example queries to provide additional clues about parameter expressions and possible values. SkillFlow-style implicitly identifies parameters from the complete skill document represented as unstructured text. Since these methods do not utilize complete structured parameter schemas, their parameter filling process mainly relies on the identified slots in generated templates and default or fallback strategies, lacking explicit constraints on parameter types, requiredness, default values, enumerated values, and value ranges.

In contrast, Skill2Query leverages parameter nodes and their attribute constraints parsed from the SKG to guide both query template generation and parameter filling. These structured constraints include parameter types, requiredness, default values, enumerated values, example values, and parameter expression patterns extracted from usage examples. During parameter instantiation, the system selects concrete values for each parameter slot according to these constraints. For the above query template, the instantiated pseudo-queries are:

\textit{How many ways can I seat 1 person at 1 identical round table?}

\textit{How many ways can I seat 5 people at 2 identical round tables?}

\subsection{Pseudo-Query Generation Quality Metrics}

We evaluate pseudo-query generation quality from three perspectives: parameter validity, functional coverage, and linguistic diversity. Specifically, we adopt Exec-Pass, Func-Coverage, and Distinct-3~\cite{li2016diversity} as evaluation metrics.

\subsubsection{Exec-Pass}

Exec-Pass measures whether the generated query templates satisfy the parameter constraints of the corresponding skills after parameter instantiation. The system checks the coverage of required parameters, parameter types, value ranges, and enumeration constraints. Given $N$ generated results, if the $i$-th result passes all parameter validation checks, we define $v_i=1$, otherwise, $v_i=0$. Exec-Pass is calculated as:

\begin{equation}
\mathrm{Exec\text{-}Pass}
=
\frac{1}{N}\sum_{i=1}^{N} v_i
\end{equation}

\subsubsection{Func-Coverage}

Func-Coverage measures the extent to which the generated query set of a skill covers its capability set. For a skill $s_i$, let its capability set be $C_i$, and let $C_i^{\mathrm{cov}}$ denote the subset of capabilities covered by at least one generated query. The functional coverage of skill $s_i$ is defined as:

\begin{equation}
FC_i=
\frac{|C_i^{\mathrm{cov}}|}
{|C_i|}
\end{equation}

The overall Func-Coverage is computed as the average functional coverage over all skills with non-empty capability sets:

\begin{equation}
\mathrm{Func\text{-}Coverage}
=
\frac{1}{|S_C|}
\sum_{s_i\in S_C} FC_i
\end{equation}

where $S_C$ denotes the set of skills for which capabilities are successfully parsed. Skills without successfully extracted capabilities are excluded from the average computation.

\subsubsection{Distinct-3}

Distinct-3 measures the linguistic diversity of the generated corpus at the trigram level. Let $G_3$ denote the multiset of all 3-grams appearing in the generated queries, and let $\mathrm{Unique}(G_3)$ denote the set of unique 3-grams. Distinct-3 is defined as:

\begin{equation}
\mathrm{Distinct\text{-}3}
=
\frac{|\mathrm{Unique}(G_3)|}
{|G_3|}
\end{equation}

A higher Distinct-3 score indicates greater diversity in lexical combinations and expression patterns. However, this metric does not directly reflect the functional correctness or parameter validity of generated queries.

\section{Implementation of the online Query Expansion System}

The online query expansion experiments use GPT-4o-mini as the generation model. The number of variants generated for each original query is set to two. In a single model call, the generator receives the original query together with the SKG-based context of the seed skills and produces two query variants for retrieval.

The prompt instructs the model to preserve the core user intent of the original query while leveraging the capabilities, parameters, examples, templates, and pseudo-queries of the candidate skills. The goal is to transform colloquial, concise, or incomplete user requests into more normalized expressions that better match the distribution of the skill retrieval corpus. To reduce incorrect guidance from candidate-skill information, the model is prohibited from introducing new tasks, tools, interfaces, concrete parameter values, or additional execution conditions that are absent from both the original query and the seed-skill context. The two generated variants are also required to remain semantically equivalent to the original query while exhibiting sufficient linguistic variation, rather than simply copying the original query or duplicating each other. The complete prompt is shown in Table~\ref{tab:online_expansion_prompt}.

\begin{table}[t]
\centering
\caption{Prompt used for online query expansion.}
\label{tab:online_expansion_prompt}
\begin{minipage}{0.95\columnwidth}
\small
\hrule
\vspace{4pt}

\textbf{System Prompt}

You are a Skill2Query online query expansion assistant.

\vspace{6pt}

\textbf{User Prompt}

You are a Skill2Query online query expansion module.

You are given an original user query and one or more candidate skills with Skill2Query-generated structured context.

Generate exactly \{num\_variants\} concise retrieval-query variants.

Generation rules:

1. Use the candidate skills' capabilities, parameters, examples, templates, and pseudo queries as grounding signals.

2. Translate the raw query into normalized skill-retrieval wording.

3. You may introduce API, tool, file-type, domain, or task terms when they come from the candidate Skill2Query context and help connect the query to a skill.

4. Do not invent details absent from both the raw query and candidate context.

5. Prefer variants that resemble the candidate templates or pseudo queries.

6. Output one variant per line with no numbering.

Original query:

\{user\_query\}

Candidate Skill2Query context:

\{context\}

Variants:

\vspace{4pt}
\hrule
\end{minipage}
\end{table}

After query expansion, the original query is retained rather than discarded. The original query and the two generated variants are combined into a query set, resulting in three query views for each test instance. Formally, the expanded query set is defined as
\[
\mathcal{Q}_{\mathrm{exp}}
=
\{q,q_1,q_2\},
\]
where $q$ is the original query and $q_1,q_2$ are the two generated
variants. Retaining the original query prevents the original semantics from being completely lost when the generated rewrites are inaccurate. It also allows the retrieval results produced by the original query to serve as a stable reference during result fusion.

For each of the three queries, the system invokes the same base retriever and performs retrieval independently over the same original skill index. Each query view returns the top 20 candidate skills, with $\mathrm{TOP\_K}=20$. The system therefore obtains three independent but semantically related ranked lists: one produced from the original query and two produced from the generated query variants.

To merge the ranked lists from multiple query views into a unified
result, the system applies unsupervised Reciprocal Rank Fusion (RRF).
For each candidate skill $s_i$, its fused score is computed as
\[
\operatorname{RRF}(s_i)
=
\sum_{q' \in \mathcal{Q}_{\mathrm{exp}}}
\frac{1}{k+\operatorname{rank}_{q'}(s_i)},
\]
where $\operatorname{rank}_{q'}(s_i)$ denotes the rank of skill $s_i$
under query view $q'$. We set the smoothing parameter to $k=60$.
All candidate skills are reranked by their RRF scores, and the fused
top-20 skills are returned as the final online retrieval results. 

The online expansion module also implements retry and fallback mechanisms for generation failures. When an LLM call fails or produces no valid query variants, the system retries the generation process up to two times. If no valid variant is obtained after the retries, the system falls back to retrieval using only the original query. In this case, the current test instance is not interrupted or discarded. And instead, the ranked results produced by the base retriever are returned directly.

This fallback mechanism ensures that the online method remains compatible with the original baseline retrieval pipeline when query expansion fails. Consequently, failures in the generation module do not interrupt evaluation or cause missing test instances, and the system can still maintain the basic retrieval procedure provided by the baseline.

\section{Downstream Skill Retrieval Training Setup}

To evaluate the effectiveness of pseudo-queries generated by different
methods in downstream skill retrieval, we fine-tune the two-stage
SkillRouter retrieval framework. The framework consists of
SkillRouter-Embedding-0.6B for first-stage dense retrieval and
SkillRouter-Reranker-0.6B for second-stage reranking. Across all
experiments, the model architectures, input formats, training
objectives, data-sampling strategies, and hyperparameter settings are
kept fixed. Only the pseudo-query generation method used
to construct the training data varies across experiments.

\subsection{Training Data Construction}

\subsubsection{Pseudo-Query Sampling}

For each pseudo-query generation method evaluated in RQ1, we pair the
generated pseudo-queries with their corresponding skills to construct
a separate training set. All training sets are processed using the same
sampling strategy and are used to fine-tune SkillRouter under an
identical training pipeline, thereby ensuring that differences in
downstream retrieval performance are primarily attributable to the
quality of the generated pseudo-queries.
During training-sample construction, at most one query template is
selected for each skill, and one pseudo-query is sampled from the
pseudo-queries associated with that template. Consequently, the
resulting training data approximately maintain a one-query-per-skill
structure. This sampling strategy prevents skills with larger numbers
of generated pseudo-queries from being overrepresented in the training
set. 

\subsubsection{Two-Stage Training Data Construction}

The first stage fine-tunes SkillRouter-Embedding-0.6B. Each training
instance consists of a pseudo-query and its corresponding positive
skill, forming a query--skill positive pair. The query is encoded with
a retrieval instruction prefix, while the skill representation is
constructed by concatenating its name, description, and body.

A unique-skill batch sampler is used during training to ensure that the
\texttt{skill\_id} values within the same batch are distinct. This
reduces the risk that training instances associated with the same skill
are incorrectly treated as in-batch negatives.

The second stage fine-tunes SkillRouter-Reranker-0.6B using query-level
candidate groups, each containing one query, one positive skill, and
multiple negative skills.

For each pseudo-query generation method, candidate skills are retrieved
using the corresponding fine-tuned embedding model and offline dense
index. The top 200 records are retrieved, deduplicated by
\texttt{skill\_id}, and reduced to the top 10 candidates.

If the positive skill is absent from the top-10 candidates, it replaces
the lowest-ranked candidate. Each group therefore contains exactly one
positive skill, while the remaining candidates are treated as negatives.

\subsection{Model Input Formats}

\subsubsection{Embedding Model Input}

The embedding model adopts a dual-encoder architecture that independently
encodes the query and the skill document. The query-side input follows
the format:

\begin{quote}
\texttt{Instruct: Given a task description, retrieve the most relevant
skill document that would help an agent complete the task}

\texttt{Query: <query text>}
\end{quote}

The skill-side input is constructed from the skill name, description,
and complete body:

\begin{quote}
\texttt{Name: <skill\_name>}

\texttt{Description: <skill\_description>}

\texttt{Body: <skill\_body>}
\end{quote}

Before tokenization, the query text is truncated to at most 1,500
characters, the skill description to at most 300 characters, and the
skill body to at most 2,500 characters. The maximum input length for
both the query encoder and the skill encoder is set to 2,048 tokens.

\subsubsection{Reranker Model Input}

The reranker jointly encodes each query and candidate skill. The
candidate skill document is also composed of its name, description, and
body, and is concatenated with the query using the following template:

\begin{quote}
\texttt{<Instruct>: Given a task description, judge whether the skill
document is relevant and useful for completing the task}

\texttt{<Query>: <query text>}

\texttt{<Skill>:}

\texttt{Name: <skill\_name>}

\texttt{Description: <skill\_description>}

\texttt{Body: <skill\_body>}
\end{quote}

Before constructing the prompt, the query text is truncated to at most
1,500 characters, the skill description to at most 500 characters, and
the skill body to at most 2,000 characters. The maximum reranker input
length is set to 4,096 tokens.

\subsection{Training Objectives}

\subsubsection{Embedding Objective}

The embedding model is trained using a unidirectional query-to-skill
in-batch InfoNCE objective~\cite{oord2018representation}. For a batch containing $B$ training
instances, the $i$-th query and the $i$-th skill form a positive pair,
while all other skills in the same batch are treated as negatives for
that query.

Let $\mathbf{q}_i$ and $\mathbf{s}_j$ denote the normalized query and
skill embeddings, respectively. The embedding loss is defined as:

\begin{equation}
\mathcal{L}_{\mathrm{emb}}
=
-\frac{1}{B}
\sum_{i=1}^{B}
\log
\frac{
\exp\left(
\operatorname{sim}(\mathbf{q}_i,\mathbf{s}_i)/\tau
\right)
}{
\sum_{j=1}^{B}
\exp\left(
\operatorname{sim}(\mathbf{q}_i,\mathbf{s}_j)/\tau
\right)
}.
\end{equation}

Here, $\operatorname{sim}(\cdot,\cdot)$ denotes the cosine similarity
between normalized embeddings, and $\tau$ is the temperature parameter,
which is set to $0.05$.

\subsubsection{Reranker Objective}

For each query-candidate pair, the reranker computes a relevance score. The relevance score is defined as the difference between the logits of
the \texttt{yes} and \texttt{no} tokens at the final output position:

\begin{equation}
f(q,s)
=
\operatorname{logit}_{\texttt{yes}}
-
\operatorname{logit}_{\texttt{no}}.
\end{equation}

For a training group containing $K$ candidate skills, let $s^{+}$
denote the unique positive skill. The reranker is trained using a
listwise cross-entropy loss:

\begin{equation}
\mathcal{L}_{\mathrm{rank}}
=
-\log
\frac{
\exp\left(f(q,s^{+})/\tau\right)
}{
\sum_{j=1}^{K}
\exp\left(f(q,s_j)/\tau\right)
}.
\end{equation}

The reranker temperature $\tau$ is set to $1.0$. This objective directly
compares the relative relevance scores of candidate skills within the
same candidate group, encouraging the positive skill to receive a higher
rank.

\subsection{Training Configuration}

The main fine-tuning configurations for the embedding and reranker
models are summarized in Table~\ref{tab:skillrouter_training_config}.
All pseudo-query generation methods use the same training
hyperparameters.

\begin{table}[t]
\centering
\caption{Fine-tuning configurations of the two-stage skill retrieval model.}
\label{tab:skillrouter_training_config}
\small
\begin{tabular}{lcc}
\toprule
Setting & SR-Emb-0.6B & SR-Re-0.6B \\
\midrule
Training objective
& In-batch InfoNCE
& Listwise cross-entropy \\

Temperature
& 0.05
& 1.0 \\

Epochs
& 1
& 1 \\

Batch size
& 8
& 1 candidate group \\

Gradient accumulation
& 1
& 16 \\

Learning rate
& $2 \times 10^{-5}$
& $1 \times 10^{-5}$ \\

Weight decay
& 0.01
& 0.01 \\

Optimizer
& AdamW
& AdamW \\

Gradient clipping
& 1.0
& 1.0 \\

Validation ratio
& 0.02
& 0.02 \\

Random seed
& 42
& 42 \\

Gradient checkpointing
& Disabled
& Enabled \\
\bottomrule
\end{tabular}
\end{table}

\section{Skill2Query Ablation Study}

\begin{figure}[t]
\centering
\includegraphics[width=\columnwidth]{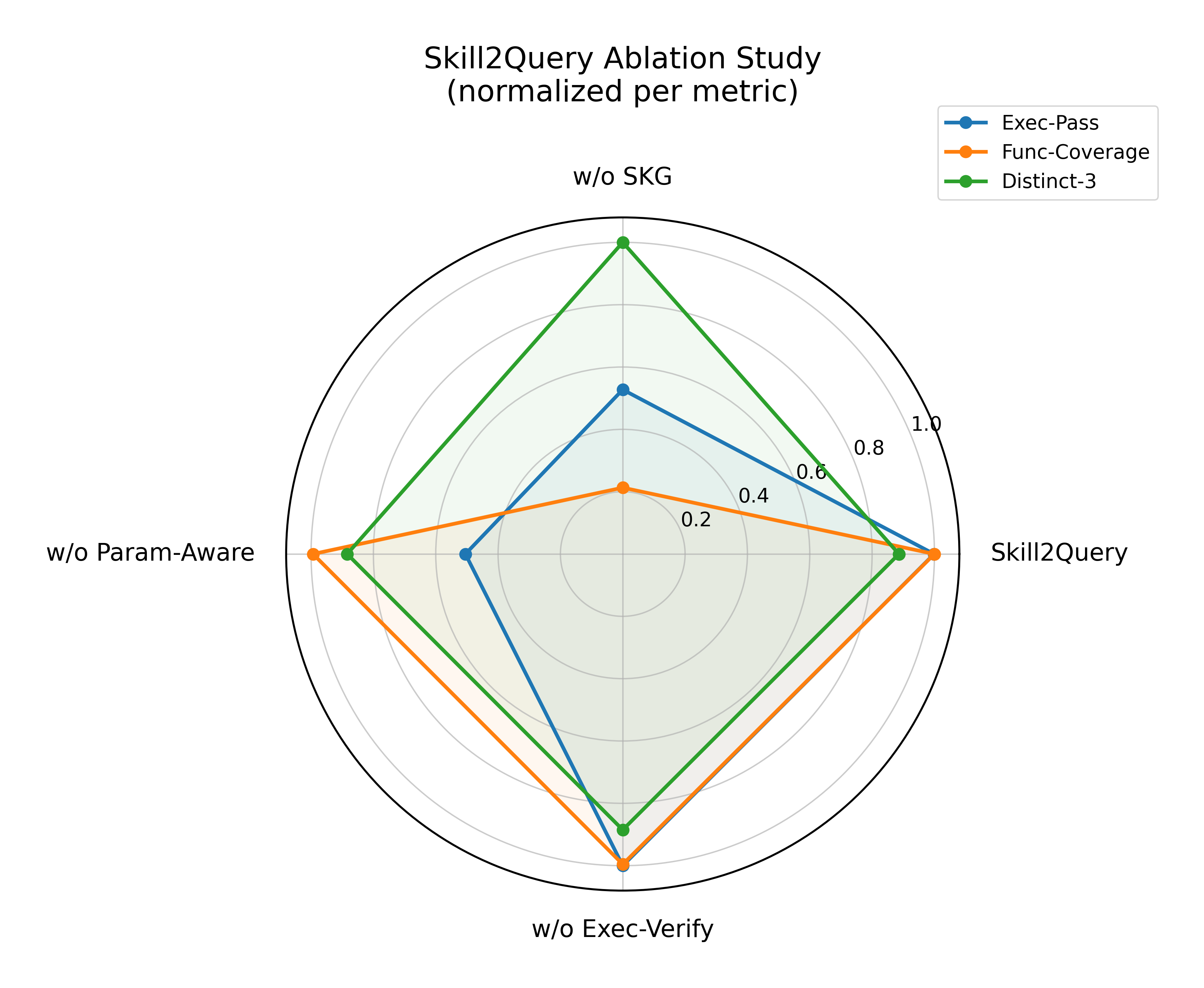}
\caption{Normalized comparison of Skill2Query and its ablation variants across the three generation-quality metrics.}
\label{fig:Skill2Query_ablation_normalized}
\end{figure}

To analyze the contribution of the core components of Skill2Query to pseudo-query generation quality, we construct three ablation variants using the complete Skill2Query framework as the reference.

w/o SKG removes the Skill Knowledge Graph and generates queries only from the plain-text skill description, allowing us to examine the contribution of structured skill representations. w/o Param-Aware removes the parameter-aware generation mechanism, such that the third layer no longer produces parameter-slot definitions and instead outputs only plain-text query templates. This variant is used to evaluate the effect of explicit parameter information on generation quality. w/o Exec-Verify removes the parameter verification process, so generated results are no longer checked for parameter names, types, value ranges, or enumeration constraints. This variant is used to assess the contribution of static verification. The original metric values are reported in Table~\ref{tab:Skill2Query_ablation},
while Figure~\ref{fig:Skill2Query_ablation_normalized} provides a normalized
comparison across the three generation-quality metrics.

\begin{table}[t]
\centering
\caption{Ablation results of Skill2Query.}
\label{tab:Skill2Query_ablation}
\small
\begin{tabular}{lccc}
\toprule
Variant
& Exec-Pass (\%)
& Func-Coverage (\%)
& Distinct-3 (\%) \\
\midrule
Skill2Query
& 42.85
& 11.32
& 85.71 \\

w/o SKG
& 22.63
& 2.41
& 96.74 \\

w/o Param-Aware
& 21.66
& 11.25
& 85.54 \\

w/o Exec-Verify
& 42.90
& 11.27
& 85.62 \\
\bottomrule
\end{tabular}
\end{table}

 Among the three components, the SKG has the most pronounced effect on functional coverage. After removing the SKG, Func-Coverage decreases from 11.32\% to 2.41\%, corresponding to a drop of 8.91 percentage points. Exec-Pass also decreases from 42.85\% to 22.63\%, a reduction of 20.22 percentage points. These results indicate that relying only on plain-text skill descriptions makes it difficult to consistently identify the distinct capabilities contained in a skill and their associated parameter constraints. In contrast, the structured organization of Capability, Parameter, and Example information in the SKG provides clearer semantic guidance for subsequent generation.

Notably, the Distinct-3 score of w/o SKG increases from 85.71\% to 96.74\%. This increase does not imply that removing the SKG improves overall generation quality. Instead, it suggests that, without structured constraints, the generated language becomes more dispersed in surface form. When considered together with the substantial decreases in Exec-Pass and Func-Coverage, this result shows that greater lexical diversity alone does not guarantee correctness in functional semantics or parameter constraints. The primary role of the SKG is therefore not to maximize surface-level variation, but to keep the generated content aligned with the target skill while maintaining a reasonable degree of linguistic diversity.

The parameter-aware generation mechanism also has a substantial effect on Exec-Pass. Removing Param-Aware reduces Exec-Pass from 42.85\% to 21.66\%, corresponding to a drop of 21.19 percentage points. In contrast, Func-Coverage decreases only slightly from 11.32\% to 11.25\%, while Distinct-3 remains nearly unchanged, decreasing from 85.71\% to 85.54\%. These results indicate that the parameter-aware mechanism primarily affects the parameter structure and executability of generated queries rather than their functional coverage or surface-level linguistic diversity. Even when plain-text query templates cover similar capabilities, the absence of explicit parameter slots and parameter-type information makes the resulting queries less likely to satisfy downstream parameter parsing and execution requirements.

In comparison, removing Exec-Verify results in only minor changes across all three metrics. Exec-Pass changes from 42.85\% to 42.90\%, Func-Coverage changes from 11.32\% to 11.27\%, and Distinct-3 changes from 85.71\% to 85.62\%. All differences are below 0.1 percentage points. This suggests that, under the current dataset and generation settings, most candidate results already satisfy the basic parameter constraints before entering the verification stage, and the static verifier therefore has limited influence on the overall average metrics.

Nevertheless, Exec-Verify still serves an important quality-control function by filtering results with invalid parameter names, incompatible parameter types, out-of-range numerical values, or illegal enumeration values. Its main contribution is therefore to prevent abnormal results from entering the final pseudo-query corpus, rather than to substantially improve average generation metrics.

Overall, the ablation results show that the SKG and parameter-aware generation mechanism play distinct but complementary roles. The SKG provides a structured semantic foundation linking capabilities, parameters, and examples, and therefore has a substantial effect on functional coverage and overall validity. Param-Aware explicitly introduces parameter structures into query templates and has the most direct effect on pseudo-query executability. Exec-Verify acts as a final quality-control module that filters the relatively small number of generated results that violate parameter constraints.
\end{document}